\documentclass{article}

\usepackage{cite}
\usepackage{authblk}

\usepackage[utf8]{inputenc} % allow utf-8 input
\usepackage[T1]{fontenc}    % use 8-bit T1 fonts
\usepackage[hidelinks]{hyperref}       % hyperlinks
\usepackage{url}            % simple URL typesetting
\usepackage{booktabs}       % professional-quality tables
\usepackage{amsfonts}       % blackboard math symbols
\usepackage{nicefrac}       % compact symbols for 1/2, etc.
\usepackage{microtype}      % microtypography
\usepackage{xcolor}         % colors
\usepackage{pifont}
\usepackage{enumitem}
\usepackage{graphicx}
\usepackage{algorithm}
\usepackage{dsfont}
\usepackage{amssymb}
\usepackage{algpseudocode}
\usepackage{subcaption}

\usepackage{ifluatex}
\ifluatex
  \usepackage{pdftexcmds}
  \makeatletter
  \let\pdfstrcmp\pdf@strcmp
  \let\pdffilemoddate\pdf@filemoddate
  \makeatother
\fi

\usepackage{multirow} 
\usepackage{svg}
\usepackage[toc,page]{appendix}
\usepackage{tabularx}
\usepackage{relsize}
\usepackage{booktabs} % Required for \toprule, \midrule, and \bottomrule
\usepackage{geometry} % Optional, for page size and margin setup
\usepackage{amsmath}
\usepackage{amsthm}
\usepackage{bbm}
\usepackage{wrapfig,lipsum,booktabs}
\newtheorem{definition}{Definition}

\newcommand{\rev}[1]{\textcolor{black}{#1}}

\newcommand{\cmark}{\ding{51}}%
\newcommand{\xmark}{\ding{55}}%

\newcommand{\Assume}[1]{\item[\textbf{Assume:}] #1}

\newtheorem{theorem}{Theorem}
\newtheorem{prop}{Proposition}
\newtheorem{lemma}{Lemma}

\usepackage{tikz}
\usetikzlibrary{positioning, shapes, arrows, calc}

\newcommand*\samethanks[1][\value{footnote}]{\footnotemark[#1]}

\title{Protected Test-Time Adaptation via Online Entropy Matching: A Betting Approach}

\author[1]{Yarin~Bar\thanks{Equal contribution.}}
\author[2]{Shalev~Shaer\samethanks}
\author[1,2]{Yaniv~Romano}

\affil[1]{Department of Computer Science, Technion IIT, Israel}
\affil[2]{Department of Electrical and Computer Engineering, Technion IIT, Israel}

\begin{document}

\date{}

\maketitle

\begin{abstract}
We present a novel approach for test-time adaptation via online self-training, consisting of two components. First, we introduce a statistical framework that detects distribution shifts in the classifier's entropy values obtained on a stream of unlabeled samples. Second, we devise an online adaptation mechanism that utilizes the evidence of distribution shifts captured by the detection tool to dynamically update the classifier's parameters. The resulting adaptation process drives the distribution of test entropy values obtained from the self-trained classifier to match those of the source domain, building invariance to distribution shifts. This approach departs from the conventional self-training method, which focuses on minimizing the classifier's entropy. Our approach combines concepts in betting martingales and online learning to form a detection tool capable of quickly reacting to distribution shifts. We then reveal a tight relation between our adaptation scheme and optimal transport, which forms the basis of our novel self-supervised loss. Experimental results demonstrate that our approach improves test-time accuracy under distribution shifts while maintaining accuracy and calibration in their absence, outperforming leading entropy minimization methods across various scenarios.

\end{abstract}

\section{Introduction}
\label{sec:intro}

The deployment of machine learning (ML) models in real-world settings presents a significant challenge, as these models often encounter testing environments (target domains) that differ from their training, source domain~\cite{Wang_2022_CVPR,varsavsky2020test,farahani2021brief,wilson2020survey,ganin2015unsupervised,patel2015visual,kouw2018introduction,peng2018visda,zhang2013domain,jain2011online,liu2022deep,blitzer2006domain,shi2022deep,QuioneroCandela2009DatasetSI,hendrycks2019benchmarking}. Consider, for example, an image recognition system employed for medical diagnostic support~\cite{valanarasu2022onthefly,ye2023multi,KARANI2021101907,he2021autoencoder,li2020community}, where the quality of images acquired during testing deviates from the training data due to factors such as equipment degradation and novel illumination conditions. ML models are sensitive to such distribution shifts, often resulting in performance deterioration, which can be unexpected~\cite{hendrycks2019benchmarking}. Ultimately, we want predictive models to dynamically adapt to new testing environments without the laborious work required to annotate new, up-to-date labels.

Recognizing this pressing need, there has been a surge in the development of adaptation methodologies to enhance test-time robustness to shifting distributions~\cite{liang2023comprehensive,Pandey_2021_CVPR,hendrycks2021faces,Zhang2021MEMOTT,hu2021mixnorm,mounsaveng2024bag,NEURIPS2021_1d497805,chen2023improved}. One commonly used approach involves jointly training a model on both the source and target domains~\cite{chen2011co,xu2022mutual,zou2018unsupervised,sun2020testtime,shen2018wasserstein,NEURIPS2021_b618c321}. However, such train-time adaptation methods assume access to unlabeled test data from the target domains, limiting the ability to adapt the model to new domains that emerge during testing. To overcome this limitation, test-time adaptation techniques offer strategies that dynamically update the model parameters as new unlabeled test points become available. In particular, leading methodologies draw inspiration from the relationship between the entropy of estimated class probabilities---a measure of confidence---and model accuracy~\cite{grandvalet2004semi,haeusser2017associative,tzeng2015simultaneous,carlucci2017autodial,saito2017asymmetric,shu2018dirtt,wang2020tent,niu2022efficient,niu2022towards}. Empirical evidence highlights that lower entropy often corresponds to higher accuracy, encouraging the development of self-supervised learning approaches that adjust model parameters by minimizing the entropy loss or the cross-entropy through the assignment of pseudo- or soft-labels to test points~\cite{10208420,chen2022contrastive,cohen2020selfsupervised,bartler2022mt3,bartler2022ttaps}.

While test-time adaptation techniques have shown promise in enhancing test accuracy under domain shifts, there is a caveat: minimizing entropy or related self-supervised loss functions \emph{without control} can lead to overconfident predictions, and may suffer from undesired, noisy model updates~\cite{zhao2023pitfalls,su2023beware,Gong2023SoTTART,press2024enigma}. In extreme cases, this approach may even cause the model to collapse and produce trivial predictions~\cite{wu2020entropy,morerio2017minimalentropy}. Indeed, without careful implementation and tuning, these techniques may not improve---or could even reduce~\cite{NEURIPS2021_b618c321}---the predictive performance in realistic settings.

In this paper, we present a novel, statistically principled approach to test-time adaptation via self-training. Our methodology is built upon two key pillars. First, we introduce an online statistical framework that monitors and detects distribution shifts in the test data influencing the models' predictions. We achieve this by sequentially testing whether the distribution of the classifier's entropy values obtained during testing deviates from the ones corresponding to the source domain. Armed with this monitoring tool, we then devise an online adaptation mechanism that leverages the accumulated evidence of distribution shifts to adaptively update model parameters. This mechanism drives the distribution of the self-trained classifier's entropy values, obtained on test data, to closely match the distribution of entropy values when applying the original model to the source domain. As a result, our proposed \textbf{Protected Online Entropy Matching} (\texttt{POEM}) method adapts the model on the fly in a controlled manner: in the absence of a distribution shift, our approach has a “no-harm” effect both on accuracy and calibration of the model, whereas under distribution shifts our experiments demonstrate an improvement of the test time accuracy, often surpassing state-of-the-art methods.

\subsection*{Contributions}
(i)~We present a sequential test for classification entropy drift detection, building on betting martingales~\cite{Shafer_2011,vovk2003testing,vovk2021protected} and online learning optimization~\cite{orabona2015scalefree,orabona2018scale,orabona2016coin} to provably attain fast reactions to shifting data. (ii)~Inspired by \cite{vovk2021protectedreg}, we show how to utilize the test martingale to analytically design a mapping function that transports the classifier entropies obtained at test time to resemble those of the source domain. Under certain assumptions, we establish connections between our online testing approach and optimal transport~\cite{villani2009optimal} as a mechanism for distribution matching. (iii)~This observation sets the foundation of the entropy-matching loss function used in \texttt{POEM}. (iv)~Numerical experiments in both continual and single-shift settings demonstrate that our approach is competitive and often outperforms strong benchmark methods that build on entropy minimization. These experiments are conducted using commonly used predictive models (ViT~\cite{dosovitskiy2021image} and ResNet~\cite{he2016deep}) on standard benchmark datasets: ImageNet-C~\cite{hendrycks_2018_2235448}, CIFAR10-C, CIFAR100-C, and OfficeHome \cite{venkateswara2017deep}.
A software package that implements our methods is available at \url{https://github.com/yarinbar/poem}.

\section{Preliminaries}

\subsection{Problem setup}
\label{sec:problem_setup}

To formalize the problem, consider a $K$-class classification problem with labeled training data ${(X_i^s,Y_i^s)}_{i=1}^n$ from a source domain, sampled i.i.d. from the source distribution $P_{XY}^s$. Here, $X^s \in \mathbb{R}^d$ represents observed covariates and $Y^s\in\{1,\ldots,K\}$ is the corresponding label. During testing, we encounter a stream of points $X^t_j$ with unknown labels $Y^t_j$, sampled from an unknown target distribution $P_{XY}^{j}$ that may shift over time $j=1,2,\dots$. To define the shifting mechanism, let $T_j: \mathbb{R}^d \rightarrow \mathbb{R}^d$ be an unknown corruption/shift function, resulting in test instances $X_j^t = T_j(X^s)$ with $X^s$ being a fresh i.i.d. sample from $P^s$. For instance, in datasets such as ImageNet-C, corruptions involve modifications such as blur or changes in illumination applied to clean source images. While such transformations alter the marginal $P_{X}^{j}$ distribution of the target domain, we assume the conditional distributions of $P_{Y \mid X}^s (Y^s \mid X^s)$ and $P_{Y|X}^j (Y_j^t| T_j(X^s))$ are the same for all $j=1,2,\dots$ \cite{courty2016optimal}.  
% the labels obtained by an oracle (e.g., a human annotator) for the clean and corrupted test points remain the same~\cite{courty2016optimal}. 

\subsection{Related work: test-time adaptation via self-training}
\label{sec:DA}
Given a pre-trained classifier $f_{\hat{\theta}}$ trained on the source domain, leading test-time adaptation approaches build on the idea of self-training to update the model parameters sequentially. Denote the adapted classifier by $f_{\hat{\theta}+\omega}$, where $\omega$ represents the modification to the parameters of the original model $\hat{\theta}$ obtained by self-training during testing. The adaptation process is typically initialized with $j=1$ and $\omega_1=\mathbf{0}$, and involves the following set of steps:
\begin{enumerate}
\item Observe a fresh test point $X^t_j$ and predict its unknown label using $f_{\hat{\theta} + \omega_{j}}(X^t_j)$.

\item Update the model parameters in a direction that reduces the self-supervised loss, i.e., $$\omega_{j+1} \leftarrow \omega_{j} - \eta \nabla_{\omega} 
\ell^{\textup{self}}\left(f_{\hat{\theta} + \omega_{j}}(X^t_j)\right),$$
where the hyper-parameter $\eta$ is the step size.
\item Set $j \leftarrow j+1$ and return to step 1.
\end{enumerate}

A common choice for $\ell^{\text{self}}$ in test-time adaptation methods is the entropy loss: 
\begin{align}
\label{eq:min_ent}
\ell^{\text{ent}}\left(f_{\hat{\theta} + \omega}(x)\right) = -\sum_{y=1}^K f_{\hat{\theta} + \omega}(x)_y \log(f_{\hat{\theta} + \omega}(x)_y),
\end{align}
where $f_{\hat{\theta} + \omega}(x)_y$ is the $y$-th entry of the classifier's softmax layer; we omit the index $j$ of $\omega$ for clarity. 

While entropy minimization has been shown to enhance test-time robustness~\cite{haeusser2017associative,tzeng2015simultaneous,carlucci2017autodial,saito2017asymmetric,shu2018dirtt,wang2020tent,niu2022efficient,niu2022towards}, it is also prone to instabilities~\cite{su2023beware,Gong2023SoTTART,zhao2023pitfalls,kim2023reliable}. For instance, enforcing $f_{\hat{\theta} + \omega}(x)_y=1$ for a fixed entry $y$ minimizes $\ell^{\text{ent}}$, but it collapses the classifier to make trivial predictions. To alleviate this, various strategies have been proposed. For example, one approach is to avoid training on samples with high entropy \cite{niu2022efficient}, as high entropy often correlates with erroneous pseudo labels. Another example is the utilization of a fallback mechanism that resets the adapted model back to the original model $f_{\hat{\theta}}$ when the average entropy becomes too small \cite{niu2022towards}. A more detailed review of test-time adaptation methods is given in Appendix~\ref{appdx:related_works}. Importantly, this line of work underscores the limitations of entropy minimization and highlights the need to better control its effect. This aligns with the goal of our work, which offers a distinct,  statistically grounded approach for test-time adaptation. While we also build on the classifier's entropy to form the adaptation mechanism, we could integrate any alternative self-supervised loss in our online distributional matching scheme.

\subsection{Testing by betting}
\label{sec:testing_exch}

A key component of our method is the proposal of an online test for distribution drift. The design of this test follows the framework of \emph{testing-by-betting} \cite{shafer2019game}.
Intuitively, one can interpret this testing framework as participating in a fair game. We begin with initial toy money, and at each time step, we observe a new test point and place a bet against the null hypothesis we aim to test. If the bet turns out to be correct, our wealth increases by the money we risked in the bet; otherwise, we lose, and our wealth decreases accordingly. Mathematically, the wealth process is formulated as a non-negative martingale, where a successful betting scheme is reflected in a growing martingale (wealth) trajectory, offering progressively stronger statistical evidence against the null hypothesis. However, if the null hypothesis is true, the game must be fair in the sense that it is unlikely to significantly grow our initial capital, no matter how sophisticated our betting strategy may be. This implies that, under the null hypothesis, it is unlikely that the martingale will grow significantly beyond its initial value.

The testing-by-betting framework is widely used in sequential settings. Notable examples include: one and two sample tests \cite{shekhar2023nonparametric,podkopaev2024sequential}, independence and conditional independence tests \cite{shaer2023model,grunwald2023anytime,podkopaev2024sequential}, exchangeability tests \cite{vovk2003testing,fedorova2012plug,vovk2022testing,duan2022interactive,ramdas2022testing,saha2024testing}, and more \cite{
shafer2019game,grunwald2020safe,shafer2011test,vovk2021values,ramdas2022game,howard2021time,waudby2023estimating,grunwald2023posterior,perez2022statistics,koolen2022log,vovk2023confidence,shekhar2023reducing,shin2022detectors,vovk2021testing,pmlr-v204-eliades23a}. This framework is also used for change-point detection and testing for uniformity \cite{vovk2020testing,dai2021testing,vovk2021retrain,vovk2021protected}, related to our drift detection problem. We draw inspiration from the protected probabilistic regression approach \cite{vovk2021protectedreg} that combines the probability integral transform and betting martingales to improve the robustness of a cumulative distribution function (CDF) estimator to distribution shift in the data. The experiments in \cite{vovk2021protectedreg} illustrate this method's ability to enhance the accuracy of a regression model, where this protection scheme assumes access to new up-to-date labels. In contrast, we focus on a completely different setup where the labels of the test points are unknown, showing how the protected regression approach can be generalized to form a self-supervised loss function. In turn, we introduce two key contributions to test-time adaptation via testing-by-betting. First, we present an adaptive online learning technique to optimize the betting mechanism. Second, we pioneer the application of testing-by-betting in this domain.

\section{Proposed method: protected online entropy matching (\texttt{POEM})}

\subsection{Preview of our method}

Let the random variable $Z^s = \ell^\text{ent}(f_{\hat{\theta}}(X^s))$ be the entropy value of the original classifier applied to a fresh sample $X^s$ from the source domain. We refer to this variable as the \emph{source entropy}.
In addition, denote by $ Z^t_j = \ell^\text{ent}(f_{\hat{\theta}+\omega_j}(X_j^t))$, $j=1,2,\ldots$,
a sequence of entropy values generated by the updated model, evaluated on a stream of unlabeled test data. We refer to $Z^t$ as the \emph{target entropy}. Our proposal uses the source and target entropies both to detect distribution shifts and adapt the model to new testing environments without relying on up-to-date labeled data. The rationale behind our method is as follows: when test data is sampled from the source distribution, there will be no deviation  between the source and target entropies, implying that there is no need to adapt the model. However, statistical deviations between the source entropies \( Z^s \) and target entropies \( Z^t \) can indicate that the model encounters test data different from the training distribution. This motivates us to introduce a self-training framework that encourages the model to generate test-time entropies \( Z^t \) that closely resemble the source entropies \( Z^s \) to build invariance to shifting data. 

To achieve this goal, we utilize the testing-by-betting approach and formulate our adaptation scheme as a game, in which we start with initial toy money and proceed as follows.
\begin{enumerate}
    \item Observe a fresh test point $X^t_j$ and use the model $f_{\hat{\theta}+\omega_j}$ to predict the label of $X^t_j$. 
    \item Compute the entropy of the classifier $Z^t_j=\ell^\text{ent}(f_{\hat{\theta}+\omega_j}(X_j^t))$.
    \item Use a betting function to place a bet against the null hypothesis that $Z_j^t$ follows the same distribution as the source entropies $Z^s$.
    \item If the bet is successful, increase the accumulated wealth process; otherwise, decrease it. 
    \item Leverage the betting function to obtain an adapted pseudo-entropy value $\tilde{Z}_j$ that better matches the distribution of $Z^s$. The intuition here is that we derive \( \tilde{Z}_j \) in a way that would reduce the toy money we would have gained if we had used the same betting strategy on $\tilde{Z}_j$.
    \item Update the model parameters: obtain  $\omega_{j+1}$ by taking a gradient step that reduces the self-supervised matching loss:\footnote{In the experiments in Section \ref{sec:exps}, we use a variation of this loss, described in Section~\ref{sec:together}.} 
    \begin{equation}
        \label{eq:match_loss}
        \ell^\text{match}({Z}^t_j,\tilde{Z}_j) = \frac{1}{2}(\ell^\text{ent}(f_{\hat{\theta}+\omega_j}(X_j^t)) - \tilde{Z}_j)^2.
    \end{equation}
    \item Update the betting strategy and return to Step 1.
\end{enumerate}

In the following sections, we describe in detail each component of the proposed adaptation scheme. Before proceeding, however, we pause to highlight the advantages of the matching loss \eqref{eq:match_loss} over entropy minimization.

\subsection{Motivating example: entropy minimization vs. entropy matching}
\label{sec:motivating_example}

\begin{figure}[ht]
  \centering
  \includegraphics[width=0.85\linewidth] {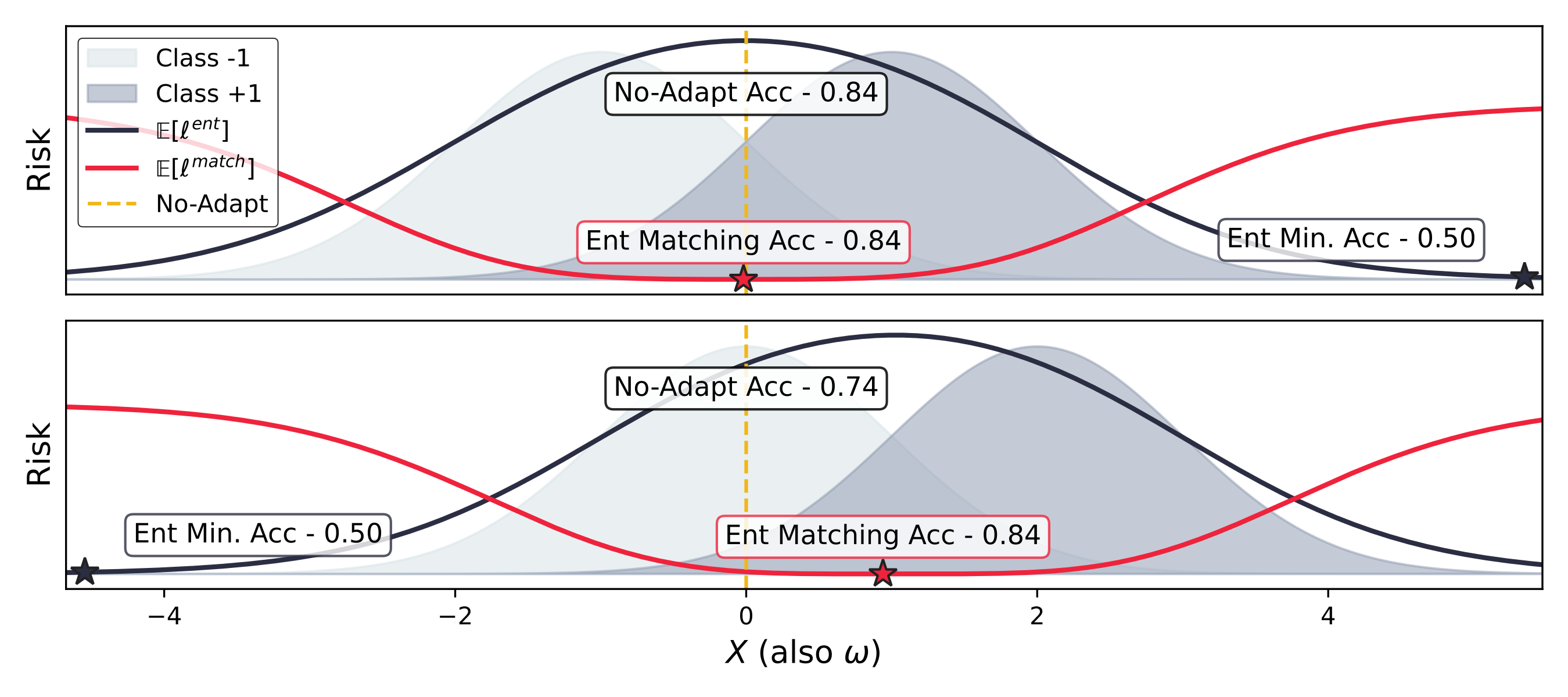}
  \vspace{-0.2cm}
  \caption{\textbf{Demonstration of the advantage of entropy matching on toy binary classification problem with Gaussian data.} The top panel represents an in-distribution setup in which $P_{XY}^t=P_{XY}^s$. The bottom panel illustrates an out-of-distribution setup, obtained by shifting the two Gaussians. The entropy matching (red) and entropy minimization (black) risks are presented as a function of $\omega$. The dashed yellow line presents the decision boundary of the pre-trained classifier. The points marked by stars correspond to the decision boundary of the adapted classifiers.
  }
  \label{fig:synth}
\end{figure}

To facilitate the exposition of the proposed loss, it is useful to consider a toy, binary classification example with a one-dimensional input $X$ in which we have oracle access both to the source $P_{XY}^s$ and a fixed target distribution $P_{XY}^t$ that does not vary over time.
We commence by generating training data 
from $P_{XY}^s$, where $P(Y=1)=P(Y=-1)$ and $P_{X \mid Y}^s = \mathcal{N}(Y^s, 1)$. 
See Figure~\ref{fig:synth} for an illustration of the source distribution. Throughout this experiment, we set the pre-trained Gaussian classifier $f_{\hat{\theta}}$ to be the Bayes optimal one for which $\hat{\theta}=0$, and during test-time we optimize the parameter $\omega$ of the updated classifier. Since $\hat{\theta}=0$, in this case $f_{\hat{\theta}+\omega}$ is simplified to $f_{\omega}$. Further implementation details are provided in Appendix~\ref{appdx:imp_details}.

As mentioned before, one of the advantages of our approach is its ``no-harm'' effect, i.e., when $P_{XY}^s=P_{XY}^t$ we ideally want to keep the decision boundary of the classifier intact. Figure~\ref{fig:synth} illustrates the entropy matching risk $\mathbb E_{X^t}[\ell^\text{match}(Z,\tilde{Z})]$ as a function of the classifier parameter $\omega$. In this synthetic case, the ideal entropy matching risk can be evaluated since we have access to the generating distribution: we can obtain the ideal pseudo-entropy $\tilde{Z}^t$, given by $\tilde{Z} = F_s^{-1}(F_t(Z^{t};f_{\omega}))$, where $F_s$ and $F_t$ are the CDF functions of the source and target entropy values, respectively; the formula above is nothing but the optimal transport map. Of course, in the practical online setting we consider in this work, $F_t$ is unknown and varies over time. In fact, this is also true in the case studied here as the distribution of $Z^t=f_{\omega}(X^t)$ varies with $\omega$, highlighting the importance of our online, adaptive testing procedure. Indeed, we see that the $\omega$ obtained by \texttt{POEM}---our online adaptation scheme---minimizes the entropy matching risk and remains close to 0, as desired.

Meanwhile, the black curve in Figure~\ref{fig:synth} illustrates the values of the entropy risk $\mathbb E_{X^t} [\ell^\text{ent}(Z^t)]$ for varying $\omega$. In contrast with our approach, the $\omega$ that minimizes the entropy risk is far from the optimal classifier. This approach results in a collapse towards a trivial classifier that always predicts $-1$, regardless of the value of $X^t$. Moving to an out-of-distribution scenario, where we consider test data sampled from a shifted version of the two Gaussians such that $Y^t=Y^s$ and $X^t=X^s+1$. 
Following the bottom panel in Figure~\ref{fig:synth}, it is evident that by minimizing the proposed entropy matching risk, the accuracy of the original (not adapted) classifier is effectively restored. In contrast, the entropy minimization paradigm once again collapses the model to make trivial predictions.

\subsection{Online drift detection}
\label{seq:shift_det}

We now turn to introduce a rigorous monitoring tool that is capable of detecting whether the distribution of the adapted classifier's entropy values ${Z}^t$, obtained at test time, deviate from $Z^s$---the source entropies obtained by applying the original model to the source data. To detect such shifts, we assume that we have access to ${F}_s$, the CDF of the source entropy values $Z^s = \ell^\text{ent}(f_{\hat{\theta}}(X^s))$. In practice, we estimate this CDF using holdout unlabeled samples $X^s$ from the source distribution. Armed with $F_s$, we then apply the probability integral transform \cite{levy_pit}, allowing us to convert any sequence of i.i.d. entropy values from the source distribution $Z^s_1, Z^s_2, \dots$ into a sequence of i.i.d. uniform random variables ${F}_s(Z^s_1), {F}_s(Z^s_2), \dots$ on the $[0,1]$ interval. Therefore, if we observe a sequence ${F}_s(Z^t_1), {F}_s(Z^t_2), \dots$ of i.i.d. uniform variables at test time, we can infer that the target entropy distribution matches the source entropy distribution. Thus, if the sequence of transformed variables ${F}_s(Z^t_1), {F}_s(Z^t_2), \dots$ deviates from the uniform distribution, we can infer that the corresponding target domain samples $Z^t_1, Z^t_2, \dots$ differ from the source distribution.
This observation lies at the core of our monitoring tool \cite{vovk2021protectedreg}.

Specifically, we leverage the testing-by-betting approach to design a sequential test for the following null hypothesis:
\begin{equation}
    \label{eq:null}
    \mathcal{H}_0: u_j \triangleq F_s(Z_j^t) \sim \mathcal{U}[0,1], \quad \forall j \in \mathbb{N},
\end{equation}
where $Z^t_j=\ell^\text{ent}(f_{\hat{\theta}+\omega_j}(X_j^t))$. In words, we continuously monitor the sequence of random variables $u_1, u_2, \ldots$ and test whether these deviate from the uniform null. We do so by formulating a test martingale, defined as follows. 
\begin{definition}[Test Martingale]
\label{def:test_martingale}
  A random process \(\{S_j : j\in \mathbb{N}, S_0=1\}\) is a test martingale for the null hypothesis \(\mathcal{H}_0\) if it satisfies the following:
  \begin{enumerate}
      \item \(S_j \ge 0 \; \ \ \forall j \in \mathbb{N}\).
      \item \(\{S_j : j\in \mathbb{N}_0\}\) is a martingale under \(\mathcal{H}_0\), i.e., $\mathbb{E}_{\mathcal{H}_0}[S_j \mid S_1, \ldots, S_{j-1}] = S_{j-1}$.
  \end{enumerate}
\end{definition}
The martingale can be thought of as the wealth process in the game-theoretical interpretation of the test, obtained by betting toy money against $\mathcal{H}_0$ as new data points arrive. We initialize this game with $S_0=1$, and update the wealth process as follows \cite{vovk2021protectedreg}: 
\begin{equation}
    \label{eq:unif_test_martingale}
    S_{j}(u_j) \triangleq S_{j-1} \cdot b(u_j) \quad \text{where} \quad b(u_j) = 1 + \epsilon_j(u_j-0.5). 
\end{equation}
Above, $b(u) \in [0,2]$ is the \emph{betting function}. The \emph{betting variable}  $\epsilon_j \in [-2,2]$ controls how aggressive the bet is, and it can be determined based on past observations $u_1, \ldots, u_{j-1}$, as we detail later in this section. 
However, before introducing our strategy to update $\epsilon_j$ over time, we should first discuss the properties of the betting function $b(u_j)$. The idea behind this choice is that we sequentially test whether the sequence of $u_1,\ldots,u_j$ observed up to time step $j$ has mean $0.5$. Indeed, if the null is true, $\mathbb{E}_{\mathcal{H}_0}[u] = 0.5$ and thus $\mathbb{E}_{\mathcal{H}_0}[b(u)] = 1$. As a result, under the null, the martingale is unlikely to grow significantly beyond its initial value---this is a consequence of Ville's inequality; see Appendix~\ref{appdx:shift_detection_with_martingale}. However, if the null is false, we can gather evidence against the uniform null hypothesis by placing more aggressive bets, especially when past observed values $u_1, \ldots, u_{j-1}$ deviate significantly from $0.5$. This highlights the role of $\epsilon_j$, controlling the value and 
direction of our wagers in each round, betting on whether $u_j$ would be over/under $0.5$. Following \eqref{eq:unif_test_martingale}, when $\epsilon_j$ and $(u_j-0.5)$ have the same sign, we win the bet and obtain $b(u_j)>1$. This implies that our capital increase as $S_{j}(u_j) := S_{j-1} \cdot b(u_j)$. Notice that, in this case, a larger $\epsilon_j$ (in absolute value) will allow us to increase the capital more rapidly, resulting in a more powerful test. However, if $\epsilon_j$ and $(u_j-0.5)$ have different signs, we lose the bet and obtain $b(u_j)<1$. Here, a larger $\epsilon_j$ would incur a more significant loss of capital. The challenge in choosing $\epsilon_j$ lies in the restriction that $\epsilon_j$ can only be determined based on past experience, i.e., we must set its value without looking at the new $u_j$. This restriction is crucial to ensure the validity of the martingale, as detailed in the following proposition.

\begin{prop}
    \label{prop:validity}
    The random process presented in~\eqref{eq:unif_test_martingale} is a valid test martingale for $\mathcal{H}_0$~\eqref{eq:null}.
\end{prop}
The proof is given in Appendix \ref{proof:prop_validity}; it is a well-known result, see, e.g., \cite{vovk2021protectedreg}.  This property is crucial to form the proposed online distributional matching mechanism, introduced in the next section. Appendix~\ref{appdx:shift_detection_with_martingale} provides further details on how the test martingale is used for distribution shift detection.

We now turn to present an adaptive approach to set the betting variable $\epsilon_j$ in a manner that enables powerful detection of drifting target entropies. This is especially important given the dynamic nature of both the target data and the continuous, online updates of the model. To achieve this, we adopt an online learning technique to learn $\epsilon_j$ from past observations, with the goal of maximizing the wealth by minimizing the negative log of the wealth process up to step $j$ \cite{shekhar2023nonparametric}:
\begin{equation}
    \label{eq:log_loss}
    -\log(S_j(u_j)) = -\log \prod_{\tau=1}^j b_\tau(u_\tau)  = -\sum_{\tau=1}^j \log(b_\tau(u_\tau)) = -\sum_{\tau=1}^j \log(1+\epsilon_\tau(u_\tau - 0.5)).
\end{equation}
This formulation allows us to learn how to predict $\epsilon_j$ using past samples via gradient descent \cite{shekhar2023nonparametric}. 

Specifically, our optimization approach relies on the scale-free online gradient descent (SF-OGD) algorithm~\cite{orabona2018scale}. Importantly, extending SF-OGD to our setting is not straightforward, since $\epsilon_j$ must be in the range of $[-2,2]$ to form a valid test martingale. In the interest of space, we present this algorithm and its theoretical analysis in Appendix~\ref{appdx:online} and only highlight here its key feature. SF-OGD allows us to attain an anytime regret guarantee, which is presented formally in Theorem~\ref{thm:sfogd_regret} of the Appendix. This guarantee bounds the difference between the negative log of the wealth process (i) obtained by the predicted $\epsilon_t$ over time horizon $1 \leq t\leq j$, and (ii) obtained by the best betting variable $\epsilon^\star$ that can only be calculated in hindsight, after looking at the data up to time $1 \leq t\leq j$. Informally, our theory shows this regret is bounded by $c \cdot \sqrt{t}$ for all $1 \leq t\leq j$, where the constant $c$ depends on the problem parameters; it is formulated precisely in Theorem~\ref{thm:sfogd_regret}. In turn, the anytime regret guarantee confirms that our SF-OGD approach effectively learns $\epsilon_j$, capturing the dynamic changes of both the target distribution and the model $f_{\hat{\theta}+\omega_j}$ in a fully online setting.

\subsection{Online model adaptation}
\label{sec:adapt}

Having established a powerful betting strategy, we now turn to show how to transform the test martingale $\{S_j : j \in \mathbb{N}\}$ into a sequence of adapted pseudo-entropy values $\tilde{Z}_1,\tilde{Z}_2,...$ that better match the distribution of $Z^s$. In what follows, we describe an algorithm to obtain $\tilde{Z}_j$, which draws inspiration from \cite{vovk2021protectedreg}, and then connect this procedure to optimal transport. 

Our adaptation scheme leverages the fact that any valid betting function is essentially a likelihood ratio~\cite{ramdas2022game, vovk2021protectedreg}. This property implies that our betting function $b(u_j) = 1 + \epsilon_j(u_j-0.5)$ satisfies
\begin{equation}
\label{eq:likelihood_ratio}
b(u_j) = \frac{dQ(u_j)}{dG(u_j)},    
\end{equation}
where $dQ(u_j)$ and $dG(u_j)$ are the densities of \emph{some} alternative distribution $Q$ and the null distribution $G$, respectively. In our case, the null distribution $G$ is the uniform distribution $\text{Uniform}(0,1)$, and the alternative distribution $Q$ can be intuitively thought of as an approximation of the unknown target entropy's CDF; we formalize this intuition hereafter. Leveraging this likelihood ratio interpretation, we follow \cite{vovk2021protectedreg} and extract the alternative distribution $Q$ by re-writing \eqref{eq:likelihood_ratio} as $dQ(u_j) = b(u_j) \cdot dG(u_j)$ and computing the integral:
\begin{align}
\label{eq:eq_Q}
Q(u_j) = \int_0^{u_j} b(v)dG(v)dv = \int_0^{u_j} b(v) \cdot 1 \cdot dv = \frac{1}{2}\epsilon_j\cdot u_j^2 + (1 - \frac{\epsilon_j}{2})\cdot u_j.    
\end{align}
Above, we used the fact that the null density is $dG(v)$ equals $1$ over the support $[0,1]$.
Having access to $Q$, we can compute the adapted $\tilde{u}_j:=Q(u_j)$ that can be intuitively interpreted as the result of applying the probability integral transform to $Z^t_j$ using the estimated target entropy CDF.
With this intuition in place, we can further convert $\tilde{u}_j$ into a pseudo-entropy value $\tilde{Z}^t_j$ that better matches the distribution of the source entropies. This is obtained by applying the inverse source CDF to $\tilde{u}_j$, resulting in $\tilde{Z}_j = F_s^{-1}(Q(u_j))$. Observe that we assume here that $F_s$ is invertible, however, in practice, we compute the pseudo-inverse of $F_s$.

To connect the adaptation scheme presented above to the optimal transport map between the target and source entropies, it is useful to consider an ideal case where we use the log-optimal bet for testing a point null \cite{ramdas2022game}.
In our case, the null hypothesis is that the distribution of the source entropies $Z^s$ and target entropies $Z^t_j$ is the same, which implies $\mathcal{H}_0$ in~\eqref{eq:null}.
Following \cite{ramdas2022game}, the optimal bet for our null is the true likelihood ratio, formulated as
\begin{equation}
\label{eq:true_likelihood_z}
    b^{\text{opt}}_Z(Z^t_j) \triangleq \frac{dF_t^j(Z^t_j)}{dF_s(Z^t_j)},
\end{equation}
where $F_t^j$ is the CDF of the target entropy $Z_j^t$. To align with \eqref{eq:likelihood_ratio}, we can equivalently write $b^{\text{opt}}_Z(Z^t_j)$ as a betting function that gets $u_j$ as an input \cite[Lemma 1]{vovk2021protectedreg}: 
\begin{equation}
\label{eq:true_likelihood_u}
    b^{\text{opt}}_Z(Z^t_j) = b^{\text{opt}}_Z(F_s^{-1}(u_j)) = \frac{dF_t^j(F_s^{-1}(u_j))}{dF_s(F_s^{-1}(u_j))} \triangleq \frac{dQ^{\text{opt}}(u_j)}{dG^{\text{opt}}(u_j)} = b^{\text{opt}}_u(u_j).
\end{equation}
Notably, this optimal betting function is infeasible to compute in practice, as $F_t^j$ is unknown. Yet, it implicitly suggests that more powerful betting functions could result in a better estimate of the target entropy CDF. Also, the optimal betting function reveals an important property of our adaptation scheme, formally given below.
\begin{prop}
    \label{prop:optimal_transport}
    Let $X_j^t$ be a fresh sample from the target domain with its corresponding $Z_j^t = \ell^\textup{ent}(f_{\hat{\theta}+\omega}(X_j^t))$ and $u_j=F_s(Z_j^t)$. Assume $F_s$ is invertible and $Z_j^t$ is continuous, and suppose the betting function represents the true likelihood ratio \eqref{eq:true_likelihood_u}. Then, the adapted $\tilde{Z}^t_j = F_s^{-1}(Q^{\textup{opt}}(u_j))$ is the optimal transport map from the target to the source entropies with respect to the Wasserstein distance.
\end{prop}
The proof of this proposition builds on \cite[Lemma 1]{vovk2021protectedreg} and is provided in Appendix \ref{proof:prop_optimal_transport}. This result highlights that our online, martingale-based adaptation scheme is grounded in optimal transport principles. This, in turn, provides a principled way to minimize the discrepancy between probability distributions. Leveraging this connection, the entropy matching loss function \eqref{eq:match_loss}, which we employ to self-train the model, can be understood as minimizing the discrepancy between the entropy distributions of the source and target domains. This implies that our loss function aligns the model's predictions across these domains. This connection also explains the ``no-harm'' effect of the proposed loss. When $P_{XY}^s=P_{XY}^{t}$ we get that $Q^{\text{opt}}(u_j)=G^{\text{opt}}(u_j)=u_j$ in the ideal case of \eqref{eq:true_likelihood_u}, implying that $\tilde{Z}_j^t=Z_j^t$. In practice, considering the betting function from \eqref{eq:unif_test_martingale}, we anticipate that $\epsilon_j$ will be close to zero thanks to our online optimization scheme, which, in turn, results in $Q(u_j) \approx u_j$ in \eqref{eq:eq_Q}.

\subsection{Putting it all together}
\label{sec:together}
Algorithm~\ref{alg:pem_main_algo} in the Appendix summarizes the entire adaptation process of \texttt{POEM}. This algorithm starts by computing the empirical CDF $\hat{F}_s$ of the source entropies to estimate $F_s$, using unlabeled holdout samples from the source domain (line~\ref{alg_line:empirical_cdf}). The betting and pseudo-entropy estimation steps are presented in lines~\ref{alg_line:place_bet}--\ref{alg_line:pseudo_inv_Fs}. Observe that in line~\ref{alg_line:pseudo_inv_Fs} we use the pseudo-inverse of $\hat{F}_s$ to obtain $\tilde{Z}_j$. The algorithm then proceeds to adapt the classifier's parameters in a direction that minimizes our self-supervised loss (line~\ref{alg_line:model_update}).  Specifically, we only update the parameters of the normalization layers $\omega$ of the classifier $f_{\hat{\theta}+\omega}$, which is a common practice in the test-time adaptation literature~\cite{wang2020tent,niu2022towards,niu2022efficient}. The self-training step is done by minimizing a variation of the entropy matching loss $\ell^{\text{match}}$ \eqref{eq:match_loss}:
\begin{equation} \label{eq:match_pp}
\ell^{\text{match++}}_{\lambda}({Z}^t_j,\tilde{Z}_j) = \ell^{\text{match}}({Z}^t_j,\tilde{Z}_j) \cdot \frac{\mathbbm{1}[{Z}^t_j < \lambda]}{\exp{\{2\cdot({Z}^t_j - \lambda})\}}, \ \ \text{where} \ \ Z_j^t=\ell^\text{ent}(f_{\hat{\theta}+\omega_j}(X_j^t)).
\end{equation}
The above loss function includes an additional sample-filtering $\mathbbm{1}[{Z}^t_j < \lambda]$ and sample-weighting $1/\exp\{{2\cdot({Z}^t_j - \lambda})\}$ components, where $\lambda > 0$ is a pre-defined thresholding parameter. The sample-filtering idea is widely used in this literature \cite{niu2022efficient,niu2022towards}, as the predictions of samples with high entropy examples tend to be inaccurate. Aligning with this intuition, the sample-weighting gives a higher weight to samples with low entropies \cite{niu2022efficient}.
Finally, we predict a new betting parameter $\epsilon_j$ for the next iteration by applying an SF-OGD step (line~\ref{alg_line:sfogd}).

\section{Experiments}
\label{sec:exps}

We conduct a comprehensive evaluation of POEM across a diverse range of datasets and scenarios commonly used in test-time adaptation literature. Our experiments span ImageNet, ImageNet-C, CIFAR10-C, and CIFAR100-C datasets for evaluating the robustness to shifts induced by corruptions, and the Office-Home dataset for domain adaptation. We study the performance of our method in both single-shift and continual-shift settings.
Our evaluation demonstrates that \texttt{POEM} is highly competitive with leading baseline test-time adaptation methods in terms of accuracy and runtime. In the interest of space, this section focuses on the results for the ImageNet dataset, as it is the most challenging one among those considered. Details and results for the experiments on CIFAR and Office-Home datasets are provided in Appendices \ref{appdx:exps_cifar} and \ref{appdx:exps_officehome}, respectively.

Throughout this section we use the test set of ImageNet to form our in-distribution dataset, and utilize ImageNet-C---which contains 15 different types of corruptions at five increasing severity levels---to simulate various out-of-distribution scenarios. \rev{Notably, since the images in ImageNet-C are variations of the same images from the ImageNet test set, our experiments simulate a realistic out-of-distribution test set by including only a single corrupted version of each image.} To demonstrate the versatility of \texttt{POEM}, we consider two pre-trained classifiers $f_{\hat{\theta}}$ of different architectures: Vision Transformer (ViT) \cite{dosovitskiy2021image} with layer norm (LN) and ResNet50 \cite{he2016deep} with group norm (GN). We compare \texttt{POEM} to four leading entropy minimization methods---\texttt{TENT}~\cite{wang2020tent}, \texttt{EATA}~\cite{niu2022efficient}, \texttt{SAR}~\cite{niu2022towards}, and \texttt{COTTA}~\cite{wang2022continual}---using code provided by the authors.
Importantly, all adaptation methods update only the normalization parameters (LN/GN) of the model, ensuring a fair comparison. Following \cite{niu2022towards}, we employ a fully online setting with a batch size of 1, in which the model is updated after observing a new test sample; see Appendix~\ref{appdx:imgnet_imp_details} for implementation details and choice of hyper-parameters. In the interest of space, we defer the results obtained by the ResNet classifier to Appendix~\ref{appdx:exps_cont} and focus here on the results obtained by the ViT model.

\paragraph{Continual shifts}\label{par:exps:continual_shift} 
\begin{figure}[ht]
  \centering
  \includegraphics[width=0.9\linewidth]{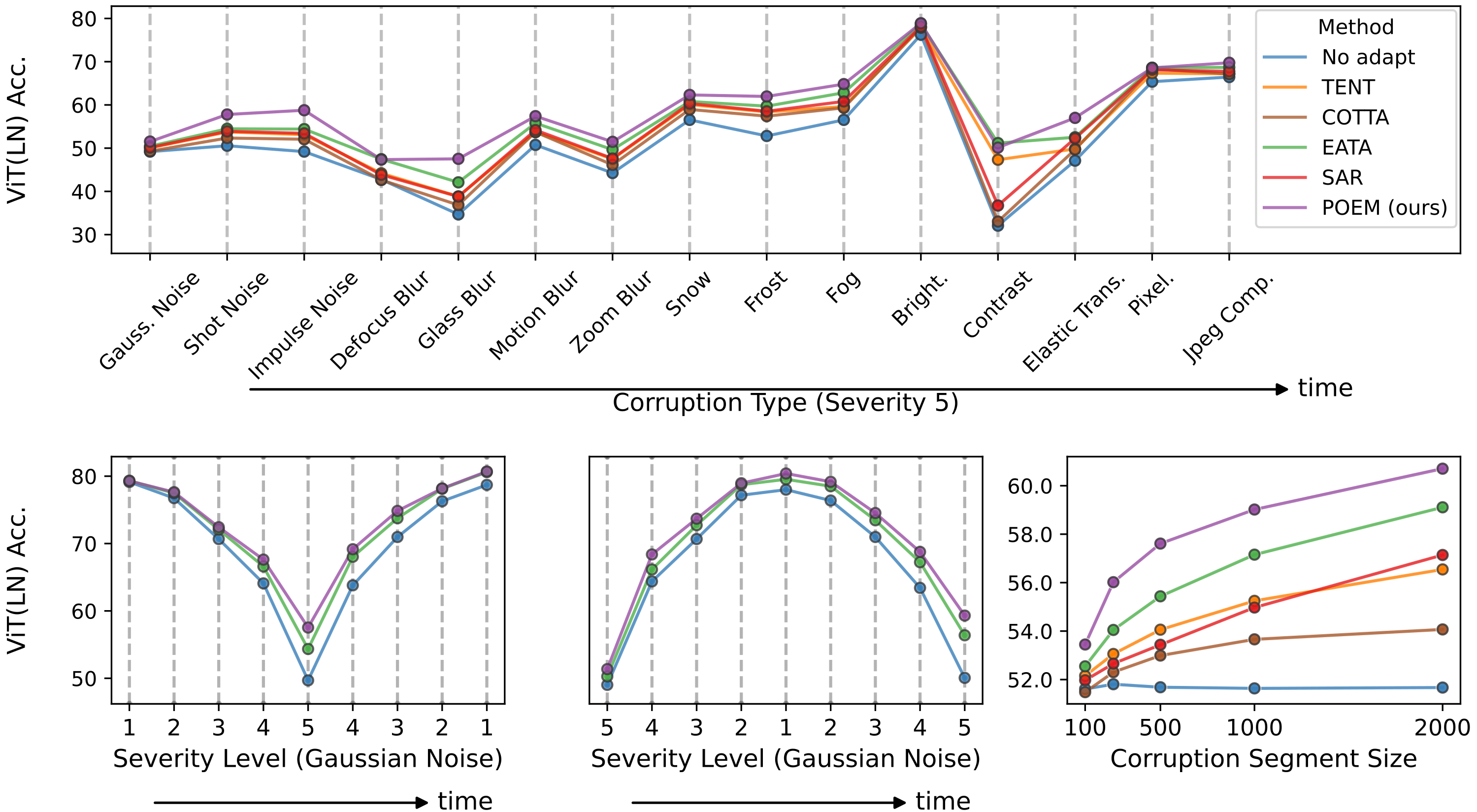}
  \vspace{-0.0cm}
  \caption{\textbf{Continual test-time adaptation on ImageNet-C with a ViT model}. \textbf{Top:} Per-corruption accuracy with a corruption segment size of 1,000 examples. Results are obtained over 10 independent trials; error bars are tiny.  \textbf{Bottom left:} Severity shift---low~(1) to high~(5) and back to low. \textbf{Bottom center:} Severity shift---high~(5) to low~(1) and back to high. To improve the readability of these two graphs, we only present \texttt{POEM}, the best-performing baseline method (\texttt{EATA}), and the \texttt{no-adapt} approach. \textbf{Bottom right:} Mean accuracy under continual corruptions as a function of the corruption segment size.}
  \label{fig:cont_main}
\end{figure}
Inspired by~\cite{wang2022continual, yuan2023robust}, we evaluate our approach in a continual setup in which sudden distribution shifts occur during testing. To simulate this, we create a test set of 15,000 samples by randomly selecting 1,000 samples from each corruption type at a fixed severity level (a corruption segment) and concatenating all 15 segments to form the test data. We apply all adaptation methods in combination with a ViT model and present the results in Figure~\ref{fig:cont_main}. Following the bottom panel in that figure, one can see that \texttt{POEM} achieves higher accuracy than all of the benchmark methods. Next, we investigate how quickly our method adapts the model to new shifts by varying the segment size of each corruption. As shown in Figure~\ref{fig:cont_main} (bottom right), \texttt{POEM} exhibits faster adaptation than the baseline methods. It successfully enhances the accuracy of the pre-trained model with as few as 100 examples per corruption (test set of size \rev{1,500}) as well as with longer adaptation using \rev{2,000} examples per corruption (test set of size \rev{30,000}).
Finally, we explore another realistic scenario of continual adaptation by varying the severity level every \rev{1,000} samples while keeping a fixed corruption type. The bottom left and center panels in Figure~\ref{fig:cont_main} show that \texttt{POEM} outperforms the best baseline method (\texttt{EATA}) in this setting as well. Lastly, similar experiments conducted with a ResNet model are presented in Figure~\ref{fig:appdx_cont_resnet} in the Appendix, showing that our method attains faster adaptation and superior accuracy than the baseline methods.

\begin{table}
    \caption{\textbf{Single shift test-time adaptation}. Accuracy evaluated on ImageNet-C, averaged over all 15 corruptions of severity level 5. Detailed results are in Table~\ref{tab:all_corruptions}. We attribute \texttt{COTTA}'s inferior performance to its learning rate being tuned for the continual setting; see Appendix~\ref{appdx:imgnet_imp_details} for details.}
    \label{tab:tta_mean_results}
    % \vspace{0.05cm}
    \centering
    \resizebox{0.3\linewidth}{!}{
    \begin{tabular}{lll}
        \toprule
         & ResNet50 & ViT  \\
        Method   &      (GN)            &    (LN)          \\
        \midrule
        No adapt &          $31.44$ &      $51.65$ \\
        \texttt{TENT}     &          $25.46$ &      $62.21$ \\
        \texttt{COTTA}    &          $23.90$ &      $55.34$ \\
        \texttt{EATA}     &          $38.63$ &      $64.14$ \\
        \texttt{SAR}      &          $36.01$ &      $64.03$ \\
        \texttt{POEM} (ours)     &          $\mathbf{38.90}$ &      $\mathbf{67.36}$ \\
        \bottomrule
    \end{tabular}}
\end{table} 
\paragraph{Single shift} We now consider a scenario with a single corruption type of a fixed severity level, which follows~\cite{wang2020tent, niu2022towards,wang2023survey}. 
Table~\ref{tab:tta_mean_results} summarizes the average results across all corruptions for severity level 5, demonstrating that \texttt{POEM} achieves an average accuracy of 67.36\% for the ViT model, outperforming the best baseline method (\texttt{EATA}) with an absolute average accuracy gap of 3.22\%.
A detailed breakdown by corruption type for each classifier is provided in Table~\ref{tab:all_corruptions} in the Appendix. Notably, \texttt{POEM} outperforms all benchmark methods on all corruption types for ViT, while achieving higher test accuracy in 9 out of the 15 corruption types for the ResNet model.

\begin{figure}[ht]
  \centering
  \includegraphics[width = \linewidth]{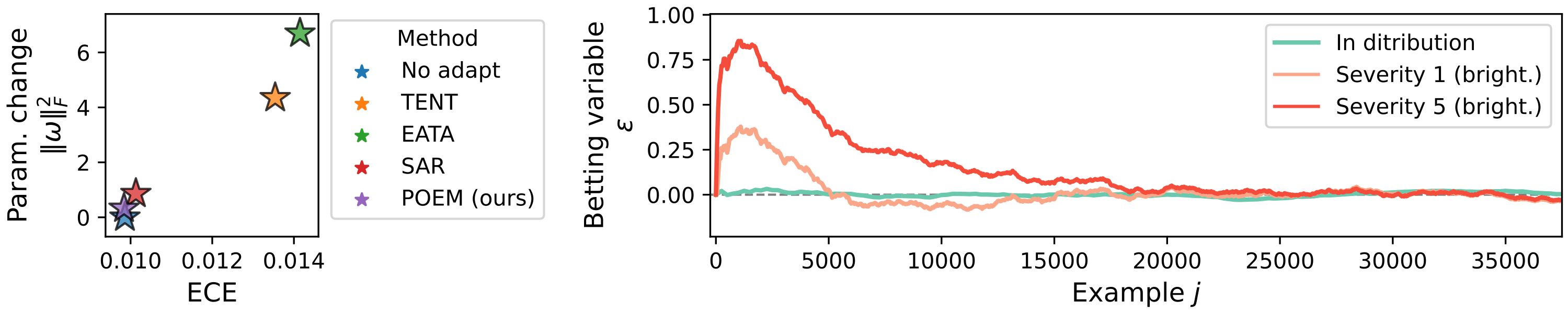}
  \vspace{-0.5cm}
  \caption{\textbf{In-distribution experiment on ImageNet (left panel)}: calibration error (ECE \cite{guo2017calibration}) versus $\|\omega\|_F^2$---a metric that evaluates the classifier's parameters deviation from the original ViT model. Lower values on both axes are better. Results are averaged across 10 independent trials; standard errors and accuracy of each method are reported in Table~\ref{tab:in_dist_performance} in the appendix.  \textbf{The behavior of the betting parameter (right panel)}: the value of $\epsilon$ is presented as a function of time for both in- and out-of-distribution experiments (a single shift, two severity levels).}
  \label{fig:in_dist_main}
\end{figure}

\paragraph{In distribution} In this setting, we apply all methods on the validation set of the ImageNet dataset. Following Table~\ref{tab:in_dist_performance} in the Appendix, all the methods maintain a similar accuracy as the original model, however, the baseline methods tend to increase the expected calibration error (ECE) \cite{kim2023reliable,guo2017calibration} and unnecessarily modify the model's parameters, as measured by $\|\omega\|^F_2$. In contrast, following Figure~\ref{fig:in_dist_main} (left panel), \texttt{POEM} exhibits minimal changes both for ECE and model parameters, as desired. Figure~\ref{fig:appdx_in_dist_resnet} in the Appendix leads to similar conclusions for the ResNet model.

\paragraph{\rev{Additional experiments on ImageNet, including ablation study}} 
The right panel of Figure~\ref{fig:in_dist_main} plots the value of the betting parameter $\epsilon$ over time, for both in- and out-of-distribution scenarios. Observe how $\epsilon$ remains near zero under the in-distribution setting, explaining the minimal change in accuracy, ECE, and model's parameters $\|\omega\|_F^2$, presented in the left panel of Figure~\ref{fig:in_dist_main}.
By contrast, when considering out-of-distribution test data of a single shift, we can see that $\epsilon$ is high at the beginning and gradually reduces over time, indicating that the self-training process adapts the model to the new environment. A similar behavior is observed for the ResNet model; see Figure~\ref{fig:appdx_in_dist_resnet} in the Appendix. This conclusion is further supported by Figure~\ref{fig:appdx_cdfs} in the Appendix, showing that the CDF of the target entropies of the adapted ResNet model nearly matches the CDF of the source entropies. In
Figure~\ref{fig:appdx_martingale} of the Appendix we show how the martingale can powerfully detect shifts---even a minor one (brightness with severity level~1)---and also show how our adaptation mechanism gradually limits
the martingale's growth. \rev{Lastly, we conduct an ablation study, comparing the test accuracy of \texttt{POEM} using two loss functions: $\ell^{\text{match}}$ from~\eqref{eq:match_loss} and $\ell^{\text{match++}}$ from~\eqref{eq:match_pp}. Table~\ref{tab:ablation_avg} in the Appendix presents these results for a single shift scenario, averaged over 15 corruptions of ImageNet-C at the highest severity level. Both loss functions improve the original model test accuracy, with $\ell^\text{match++}$ showing a boost in the adaptation performance.}

\paragraph{\rev{Experiments on CIFAR-10C, CIFAR100-C, and OfficeHome datasets, sensitivity study, and runtime comparisons}} \rev{Appendices \ref{appdx:exps_cifar} and \ref{appdx:exps_officehome} present experiments on these additional datasets, leading to similar conclusions about the competitive adaptation accuracy of \texttt{POEM} compared to baseline methods. Notably, Appendix \ref{appdx:exps_cifar} includes experiments on both relatively short and long adaptation streams, with lengths of 15,000 and 112,500 samples, respectively. These experiments also include a sensitivity study on the learning rate $\eta$ used for self-training the model, showing that \texttt{POEM} exhibits greater stability across different learning rate values compared to \texttt{SAR} and \texttt{TENT}. Additionally, these experiments highlight that \texttt{POEM}'s runtime is comparable to \texttt{TENT} and \texttt{EATA}, and faster than \texttt{SAR}.}

\section{Discussion}
\label{sec:discussion_and_impact}
\rev{We introduced a novel, martingale-based approach for test-time adaptation that drives the test-time entropies of the self-trained model to match the distribution of source entropies. We validated our approach with numerical experiments, demonstrating that: (i) under in-distribution settings, \texttt{POEM} maintains the performance of the source model while avoiding overconfident predictions, a crucial advantage over entropy minimization methods; (ii) in relatively short out-of distribution test periods, our approach achieves faster adaptation than entropy minimization methods, which is attributed to our betting scheme that quickly reacts to distribution shifts; and (iii) in extended test periods, \texttt{POEM} achieves comparable adaptation performance and stability to strong baseline methods. In contrast to our approach, these baseline methods employ different strategies to stabilize long-term adaptation, such as resetting the self-trained model to its original state under specific heuristic conditions (as in \texttt{SAR}) or incorporating an anti-forgetting component into the entropy loss (as in \texttt{EATA}). Notably, integrating such stabilizing components into our method could further enhance its robustness in long-range adaptation settings. A unique feature of our monitoring tool is that it can be used to make rigorous decisions on model-resetting, for example, to prevent aggressive adaptation from a diverged state.}

\rev{One limitation of our method is the requirement of holdout unlabeled data from the source domain, used to estimate the source CDF. However, this CDF is pre-computed offline, and at test time we do not require any additional access to samples from the source data. In that respect, our work does not differ significantly from the requirements of \texttt{EATA}, which also assumes access to holdout source examples. Another limitation of our approach is the choice of its hyperparameters, particularly the learning rate $\eta$ for self-training. However, our sensitivity study showed that our method is fairly robust to this choice, especially compared to baseline methods. Lastly, similar to other experimental works, we anticipate that \texttt{POEM} may fail to improve accuracy in other settings that we have not explored, especially when facing an aggressive shift. Yet, our monitoring tool can be used to alert when facing distribution shifts, which is an important mechanism that does not appear in other test-time adaptation methods.}

\rev{In future work, we plan to complement our empirical findings with a theoretical analysis. Our goal is to rigorously determine when entropy matching is superior to entropy minimization, thereby uncovering the theoretical benefits and limitations of our approach.}

\rev{Another future direction is to support \texttt{POEM} with the ability to handle label shift at test time. This challenge is exemplified by scenarios where the source domain has a balanced label distribution, but the test domain becomes unbalanced. In such cases, our current monitoring tool might detect this label shift and trigger unnecessary adaptation in the absence of covariate shift. This underscores the need for a monitoring tool that remains invariant to label shifts.
To address this challenge, one may consider two potential approaches. The first builds on ideas from \cite{gong2022note}, particularly their prediction-balanced reservoir sampling technique. This method can be used to approximately simulate an i.i.d. data stream from a non-i.i.d. stream in a class-balanced manner, potentially reducing our martingale process's sensitivity to label shifts.
The second approach may involve the use of a weighted source CDF instead of the standard source CDF, with weights corresponding to the likelihood ratio $P^t(Y)/P^s(Y)$. This concept, borrowed from conformal prediction literature \cite{podkopaev2021distribution}, aims to make the test loss to ``look exchangeable'' with the source losses, thus adjusting for label shift. The main challenge here lies in reasonably approximating the likelihood ratio $P^t(Y)/P^s(Y)$, especially when facing simultaneous covariate and label shifts at test time. The ideas presented in \cite{zhou2023ods} may offer a promising starting point for exploring this avenue.}

\rev{More broadly, our paper introduces a rigorous online mechanism for matching the distributions of any given self-supervised loss values between source and target data. While our focus was on the entropy loss, it would be insightful to explore how our matching paradigm performs with other self-supervised loss functions, such as the soft likelihood ratio used in \cite{marsden2024universal}, and other approaches \cite{lu2023metatsallisentropy, liu2021cycle, goyal2022testtime}. We believe such explorations could potentially advance the state-of-the-art in test-time adaptation.}

\bibliographystyle{unsrt}
\bibliography{bibliography}

\begin{thebibliography}{100}

\bibitem{Wang_2022_CVPR}
Qin Wang, Olga Fink, Luc Van~Gool, and Dengxin Dai.
\newblock Continual test-time domain adaptation.
\newblock In {\em Proceedings of the IEEE/CVF Conference on Computer Vision and Pattern Recognition}, pages 7201--7211, June 2022.

\bibitem{varsavsky2020test}
Thomas Varsavsky, Mauricio Orbes-Arteaga, Carole~H Sudre, Mark~S Graham, Parashkev Nachev, and M~Jorge Cardoso.
\newblock Test-time unsupervised domain adaptation.
\newblock In {\em International Conference in Medical Image Computing and Computer Assisted Intervention}, pages 428--436. Springer, 2020.

\bibitem{farahani2021brief}
Abolfazl Farahani, Sahar Voghoei, Khaled Rasheed, and Hamid~R Arabnia.
\newblock A brief review of domain adaptation.
\newblock {\em Advances in data science and information engineering}, pages 877--894, 2021.

\bibitem{wilson2020survey}
Garrett Wilson and Diane~J Cook.
\newblock A survey of unsupervised deep domain adaptation.
\newblock {\em ACM Transactions on Intelligent Systems and Technology (TIST)}, 11(5):pp. 1--46, 2020.

\bibitem{ganin2015unsupervised}
Yaroslav Ganin and Victor Lempitsky.
\newblock Unsupervised domain adaptation by backpropagation.
\newblock In {\em Proceedings of the International Conference on Machine Learning}, pages 1180--1189. PMLR, 2015.

\bibitem{patel2015visual}
Vishal~M Patel, Raghuraman Gopalan, Ruonan Li, and Rama Chellappa.
\newblock Visual domain adaptation: A survey of recent advances.
\newblock {\em IEEE signal processing magazine}, 32(3):pp. 53--69, 2015.

\bibitem{kouw2018introduction}
Wouter~M Kouw and Marco Loog.
\newblock An introduction to domain adaptation and transfer learning.
\newblock {\em arXiv preprint arXiv:1812.11806}, 2018.

\bibitem{peng2018visda}
Xingchao Peng, Ben Usman, Neela Kaushik, Dequan Wang, Judy Hoffman, and Kate Saenko.
\newblock Visda: A synthetic-to-real benchmark for visual domain adaptation.
\newblock In {\em Proceedings of the IEEE Conference on Computer Vision and Pattern Recognition Workshops}, pages 2021--2026, 2018.

\bibitem{zhang2013domain}
Kun Zhang, Bernhard Sch{\"o}lkopf, Krikamol Muandet, and Zhikun Wang.
\newblock Domain adaptation under target and conditional shift.
\newblock In {\em Proceedings of the International Conference on Machine Learning}, pages 819--827. PMLR, 2013.

\bibitem{jain2011online}
Vidit Jain and Erik Learned-Miller.
\newblock Online domain adaptation of a pre-trained cascade of classifiers.
\newblock In {\em Proceedings of the IEEE Conference on Computer Vision and Pattern Recognition}, pages 577--584. IEEE, 2011.

\bibitem{liu2022deep}
Xiaofeng Liu, Chaehwa Yoo, Fangxu Xing, Hyejin Oh, Georges~El Fakhri, Je-Won Kang, and Jonghye Woo.
\newblock Deep unsupervised domain adaptation: A review of recent advances and perspectives.
\newblock {\em APSIPA Transactions on Signal and Information Processing}, 11, 2022.

\bibitem{blitzer2006domain}
John Blitzer, Ryan McDonald, and Fernando Pereira.
\newblock Domain adaptation with structural correspondence learning.
\newblock In {\em Proceedings Conference on Empirical Methods in Natural Language Processing}, pages 120--128, 2006.

\bibitem{shi2022deep}
Yongjie Shi, Xianghua Ying, and Jinfa Yang.
\newblock Deep unsupervised domain adaptation with time series sensor data: A survey.
\newblock {\em Sensors}, 22(15), 2022.

\bibitem{QuioneroCandela2009DatasetSI}
Joaquin Quionero-Candela, Masashi Sugiyama, Anton Schwaighofer, and Neil~D. Lawrence.
\newblock {\em Dataset Shift in Machine Learning}.
\newblock The MIT Press, 2009.

\bibitem{hendrycks2019benchmarking}
Dan Hendrycks and Thomas Dietterich.
\newblock Benchmarking neural network robustness to common corruptions and perturbations.
\newblock In {\em International Conference on Learning Representations}, 2018.

\bibitem{valanarasu2022onthefly}
Jeya Maria~Jose Valanarasu, Pengfei Guo, Vibashan VS, and Vishal~M. Patel.
\newblock On-the-fly test-time adaptation for medical image segmentation.
\newblock {\em arXiv preprint arXiv:2203.05574}, 2022.

\bibitem{ye2023multi}
Yanyu Ye, Zhenxi Zhang, Wei Wei, and Chunna Tian.
\newblock Multi task consistency guided source-free test-time domain adaptation medical image segmentation.
\newblock {\em arXiv preprint arXiv:2310.11766}, 2023.

\bibitem{KARANI2021101907}
Neerav Karani, Ertunc Erdil, Krishna Chaitanya, and Ender Konukoglu.
\newblock Test-time adaptable neural networks for robust medical image segmentation.
\newblock {\em Medical Image Analysis}, 68:101907, 2021.

\bibitem{he2021autoencoder}
Yufan He, Aaron Carass, Lianrui Zuo, Blake~E Dewey, and Jerry~L Prince.
\newblock Autoencoder based self-supervised test-time adaptation for medical image analysis.
\newblock {\em Medical Image Analysis}, 72:102136, 2021.

\bibitem{li2020community}
Zhang Li, Zheng Zhong, Yang Li, Tianyu Zhang, Liangxin Gao, Dakai Jin, Yue Sun, Xianghua Ye, Li~Yu, Zheyu Hu, Jing Xiao, Lingyun Huang, and Yuling Tang.
\newblock From community-acquired pneumonia to {COVID-19}: a deep learning--based method for quantitative analysis of {COVID-19} on thick-section {CT} scans.
\newblock {\em European Radiology}, 30(12):6828--6837, 2020.

\bibitem{liang2023comprehensive}
Jian Liang, Ran He, and Tieniu Tan.
\newblock A comprehensive survey on test-time adaptation under distribution shifts.
\newblock {\em arXiv preprint arXiv:2303.15361}, 2023.

\bibitem{Pandey_2021_CVPR}
Prashant Pandey, Mrigank Raman, Sumanth Varambally, and Prathosh AP.
\newblock Generalization on unseen domains via inference-time label-preserving target projections.
\newblock In {\em Proceedings of the IEEE/CVF Conference on Computer Vision and Pattern Recognition}, pages 12924--12933, June 2021.

\bibitem{hendrycks2021faces}
Dan Hendrycks, Steven Basart, Norman Mu, Saurav Kadavath, Frank Wang, Evan Dorundo, Rahul Desai, Tyler Zhu, Samyak Parajuli, Mike Guo, Dawn Song, Jacob Steinhardt, and Justin Gilmer.
\newblock The many faces of robustness: A critical analysis of out-of-distribution generalization.
\newblock {\em ICCV}, 2021.

\bibitem{Zhang2021MEMOTT}
Marvin Zhang, Sergey Levine, and Chelsea Finn.
\newblock {MEMO}: Test time robustness via adaptation and augmentation.
\newblock {\em Advances in neural information processing systems}, 35:38629--38642, 2022.

\bibitem{hu2021mixnorm}
Xuefeng Hu, Gokhan Uzunbas, Sirius Chen, Rui Wang, Ashish Shah, Ram Nevatia, and Ser-Nam Lim.
\newblock Mixnorm: Test-time adaptation through online normalization estimation.
\newblock {\em arXiv preprint arXiv:2110.11478}, 2021.

\bibitem{mounsaveng2024bag}
Saypraseuth Mounsaveng, Florent Chiaroni, Malik Boudiaf, Marco Pedersoli, and Ismail Ben~Ayed.
\newblock Bag of tricks for fully test-time adaptation.
\newblock In {\em Proceedings of the IEEE/CVF Winter Conference on Applications of Computer Vision}, pages 1936--1945, 2024.

\bibitem{NEURIPS2021_1d497805}
Eric Mintun, Alexander Kirillov, and Saining Xie.
\newblock On interaction between augmentations and corruptions in natural corruption robustness.
\newblock {\em Advances in Neural Information Processing Systems}, 34:3571--3583, 2021.

\bibitem{chen2023improved}
Liang Chen, Yong Zhang, Yibing Song, Ying Shan, and Lingqiao Liu.
\newblock Improved test-time adaptation for domain generalization.
\newblock In {\em Proceedings of the IEEE/CVF Conference on Computer Vision and Pattern Recognition}, pages 24172--24182, 2023.

\bibitem{chen2011co}
Minmin Chen, Kilian~Q Weinberger, and John Blitzer.
\newblock Co-training for domain adaptation.
\newblock {\em Advances in neural information processing systems}, 24:2456--2464, 2011.

\bibitem{xu2022mutual}
Yuanyuan Xu, Meina Kan, Shiguang Shan, and Xilin Chen.
\newblock Mutual learning of joint and separate domain alignments for multi-source domain adaptation.
\newblock In {\em WACV}, pages 1890--1899, 2022.

\bibitem{zou2018unsupervised}
Yang Zou, Zhiding Yu, BVK Kumar, and Jinsong Wang.
\newblock Unsupervised domain adaptation for semantic segmentation via class-balanced self-training.
\newblock In {\em Proceedings of the European conference on computer vision (ECCV)}, pages 289--305, 2018.

\bibitem{sun2020testtime}
Yu~Sun, Xiaolong Wang, Zhuang Liu, John Miller, Alexei Efros, and Moritz Hardt.
\newblock Test-time training with self-supervision for generalization under distribution shifts.
\newblock In {\em Proceedings of the International Conference on Machine Learning}, pages 9229--9248. PMLR, 2020.

\bibitem{shen2018wasserstein}
Jian Shen, Yanru Qu, Weinan Zhang, and Yong Yu.
\newblock Wasserstein distance guided representation learning for domain adaptation.
\newblock In {\em Proceedings of the AAAI conference on artificial intelligence}, volume~32, 2018.

\bibitem{NEURIPS2021_b618c321}
Yuejiang Liu, Parth Kothari, Bastien van Delft, Baptiste Bellot-Gurlet, Taylor Mordan, and Alexandre Alahi.
\newblock {TTT++}: When does self-supervised test-time training fail or thrive?
\newblock {\em Advances in neural information processing systems}, 2021.

\bibitem{grandvalet2004semi}
Yves Grandvalet and Yoshua Bengio.
\newblock Semi-supervised learning by entropy minimization.
\newblock {\em Advances in neural information processing systems}, 17, 2004.

\bibitem{haeusser2017associative}
Philip Haeusser, Thomas Frerix, Alexander Mordvintsev, and Daniel Cremers.
\newblock Associative domain adaptation.
\newblock In {\em Proceedings of the IEEE international conference on computer vision}, pages 2765--2773, 2017.

\bibitem{tzeng2015simultaneous}
Eric Tzeng, Judy Hoffman, Trevor Darrell, and Kate Saenko.
\newblock Simultaneous deep transfer across domains and tasks.
\newblock In {\em Proceedings of the IEEE International Conference on Computer Vision (ICCV)}, pages 4068--4076, 2015.

\bibitem{carlucci2017autodial}
Fabio~Maria Carlucci, Lorenzo Porzi, Barbara Caputo, Elisa Ricci, and Samuel~Rota Bulo.
\newblock {AutoDIAL}: Automatic domain alignment layers.
\newblock In {\em Proceedings of the IEEE International Conference on Computer Vision (ICCV)}, pages 5077--5085, 2017.

\bibitem{saito2017asymmetric}
Kuniaki Saito, Yoshitaka Ushiku, and Tatsuya Harada.
\newblock Asymmetric tri-training for unsupervised domain adaptation.
\newblock In {\em Proceedings of the International Conference on Machine Learning}, pages 2988--2997, 2017.

\bibitem{shu2018dirtt}
Rui Shu, Hung~H Bui, Hirokazu Narui, and Stefano Ermon.
\newblock A {DIRT-T} approach to unsupervised domain adaptation.
\newblock In {\em Proceedings of the International Conference on Learning Representations}, 2018.

\bibitem{wang2020tent}
Dequan Wang, Evan Shelhamer, Shaoteng Liu, Bruno Olshausen, and Trevor Darrell.
\newblock Tent: Fully test-time adaptation by entropy minimization.
\newblock In {\em International Conference on Learning Representations}, 2020.

\bibitem{niu2022efficient}
Shuaicheng Niu, Jiaxiang Wu, Yifan Zhang, Yaofo Chen, Shijian Zheng, Peilin Zhao, and Mingkui Tan.
\newblock Efficient test-time model adaptation without forgetting.
\newblock In {\em Proceedings of the International Conference on Machine Learning}, pages 16888--16905, 2022.

\bibitem{niu2022towards}
Shuaicheng Niu, Jiaxiang Wu, Yifan Zhang, Zhiquan Wen, Yaofo Chen, Peilin Zhao, and Mingkui Tan.
\newblock Towards stable test-time adaptation in dynamic wild world.
\newblock In {\em International Conference on Learning Representations}, 2023.

\bibitem{10208420}
Masud An-Nur Islam~Fahim and Jani Boutellier.
\newblock {SS-TTA}: Test-time adaption for self-supervised denoising methods.
\newblock In {\em IEEE/CVF Conference on Computer Vision and Pattern Recognition Workshops (CVPRW)}, pages 1178--1187, 2023.

\bibitem{chen2022contrastive}
Dian Chen, Dequan Wang, Trevor Darrell, and Sayna Ebrahimi.
\newblock Contrastive test-time adaptation.
\newblock In {\em Proceedings of the IEEE/CVF Conference on Computer Vision and Pattern Recognition}, pages 295--305, 2022.

\bibitem{cohen2020selfsupervised}
Tomer Cohen, Noy Shulman, Hai Morgenstern, Roey Mechrez, and Erez Farhan.
\newblock Self-supervised dynamic networks for covariate shift robustness.
\newblock {\em arXiv preprint arXiv:2006.03952}, 2020.

\bibitem{bartler2022mt3}
Alexander Bartler, Andre B{\"u}hler, Felix Wiewel, Mario D{\"o}bler, and Bin Yang.
\newblock Mt3: Meta test-time training for self-supervised test-time adaption.
\newblock {\em International Conference on Artificial Intelligence and Statistics}, pages 3080--3090, 2022.

\bibitem{bartler2022ttaps}
Alexander Bartler, Florian Bender, Felix Wiewel, and Bin Yang.
\newblock {TTAPS}: Test-time adaption by aligning prototypes using self-supervision.
\newblock In {\em IEEE International Joint Conference on Neural Networks (IJCNN)}, pages 1--8, 2022.

\bibitem{zhao2023pitfalls}
Hao Zhao, Yuejiang Liu, Alexandre Alahi, and Tao Lin.
\newblock On pitfalls of test-time adaptation.
\newblock In {\em International Conference on Machine Learning (ICML)}, 2023.

\bibitem{su2023beware}
Yi~Su, Yixin Ji, Juntao Li, Hai Ye, and Min Zhang.
\newblock Beware of model collapse! fast and stable test-time adaptation for robust question answering.
\newblock In {\em Conference on Empirical Methods in Natural Language Processing}, 2023.

\bibitem{Gong2023SoTTART}
Taesik Gong, Yewon Kim, Taeckyung Lee, Sorn Chottananurak, and Sung-Ju Lee.
\newblock {SoTTA}: Robust test-time adaptation on noisy data streams.
\newblock {\em Advances in Neural Information Processing Systems}, 36, 2024.

\bibitem{press2024enigma}
Ori Press, Ravid Shwartz-Ziv, Yann LeCun, and Matthias Bethge.
\newblock The entropy enigma: Success and failure of entropy minimization.
\newblock {\em arXiv preprint arXiv:2405.05012}, 2024.

\bibitem{wu2020entropy}
Xiaofu Wu, Suofei Zhang, Quan Zhou, Zhen Yang, Chunming Zhao, and Longin~Jan Latecki.
\newblock Entropy minimization versus diversity maximization for domain adaptation.
\newblock {\em IEEE Transactions on Neural Networks and Learning Systems}, 34(6):2896--2907, 2021.

\bibitem{morerio2017minimalentropy}
Pietro Morerio, Jacopo Cavazza, and Vittorio Murino.
\newblock Minimal-entropy correlation alignment for unsupervised deep domain adaptation.
\newblock In {\em International Conference on Learning Representations}, 2018.

\bibitem{Shafer_2011}
Glenn Shafer, Alexander Shen, Nikolai Vereshchagin, and Vladimir Vovk.
\newblock Test martingales, bayes factors and p-values.
\newblock {\em Statistical Science}, 26(1), February 2011.

\bibitem{vovk2003testing}
Vladimir Vovk, Ilia Nouretdinov, and Alexander Gammerman.
\newblock Testing exchangeability on-line.
\newblock In {\em Proceedings of the 20th International Conference on Machine Learning (ICML)}, pages 768--775, 2003.

\bibitem{vovk2021protected}
Vladimir Vovk, Ivan Petej, and Alex Gammerman.
\newblock Protected probabilistic classification.
\newblock In {\em Conformal and Probabilistic Prediction and Applications}, pages 297--299, 2021.

\bibitem{orabona2015scalefree}
Francesco Orabona and D{\'a}vid P{\'a}l.
\newblock Scale-free algorithms for online linear optimization.
\newblock In {\em International Conference on Algorithmic Learning Theory}, pages 287--301, 2015.

\bibitem{orabona2018scale}
Francesco Orabona and D{\'a}vid P{\'a}l.
\newblock Scale-free online learning.
\newblock {\em Theoretical Computer Science}, 716:50--69, 2018.

\bibitem{orabona2016coin}
Francesco Orabona and D{\'a}vid P{\'a}l.
\newblock Coin betting and parameter-free online learning.
\newblock {\em Advances in Neural Information Processing Systems}, 29:577--585, 2016.

\bibitem{vovk2021protectedreg}
Vladimir Vovk.
\newblock Protected probabilistic regression.
\newblock Technical report, Tech. Rep, 2021.

\bibitem{villani2009optimal}
C{\'e}dric Villani.
\newblock {\em Optimal transport: old and new}, volume 338.
\newblock Springer, 2009.

\bibitem{dosovitskiy2021image}
Alexey Dosovitskiy, Lucas Beyer, Alexander Kolesnikov, Dirk Weissenborn, Xiaohua Zhai, Thomas Unterthiner, Mostafa Dehghani, Matthias Minderer, Georg Heigold, Sylvain Gelly, et~al.
\newblock An image is worth 16x16 words: Transformers for image recognition at scale.
\newblock In {\em International Conference on Learning Representations}, 2020.

\bibitem{he2016deep}
Kaiming He, Xiangyu Zhang, Shaoqing Ren, and Jian Sun.
\newblock Deep residual learning for image recognition.
\newblock In {\em Proceedings of the IEEE/CVF Conference on Computer Vision and Pattern Recognition}, pages 770--778, 2016.

\bibitem{hendrycks_2018_2235448}
Dan Hendrycks and Thomas Dietterich.
\newblock Benchmarking neural network robustness to common corruptions and perturbations.
\newblock In {\em International Conference on Learning Representations}, 2018.

\bibitem{venkateswara2017deep}
Hemanth Venkateswara, Jose Eusebio, Shayok Chakraborty, and Sethuraman Panchanathan.
\newblock Deep hashing network for unsupervised domain adaptation.
\newblock In {\em Proceedings of the IEEE conference on computer vision and pattern recognition}, pages 5018--5027, 2017.

\bibitem{courty2016optimal}
Nicolas Courty, R{\'e}mi Flamary, Devis Tuia, and Alain Rakotomamonjy.
\newblock Optimal transport for domain adaptation.
\newblock {\em IEEE transactions on pattern analysis and machine intelligence}, 39(9):1853--1865, 2016.

\bibitem{kim2023reliable}
Eungyeup Kim, Mingjie Sun, Aditi Raghunathan, and J~Zico Kolter.
\newblock Reliable test-time adaptation via agreement-on-the-line.
\newblock In {\em NeurIPS Workshop on Distribution Shifts: New Frontiers with Foundation Models}, 2023.

\bibitem{shafer2019game}
Glenn Shafer and Vladimir Vovk.
\newblock Game-theoretic foundations for probability and finance.
\newblock {\em Wiley Series in Probability and Statistics}, 2019.

\bibitem{shekhar2023nonparametric}
Shubhanshu Shekhar and Aaditya Ramdas.
\newblock Nonparametric two-sample testing by betting.
\newblock {\em IEEE Transactions on Information Theory}, 2023.

\bibitem{podkopaev2024sequential}
Aleksandr Podkopaev and Aaditya Ramdas.
\newblock Sequential predictive two-sample and independence testing.
\newblock {\em Advances in Neural Information Processing Systems}, 36, 2024.

\bibitem{shaer2023model}
Shalev Shaer, Gal Maman, and Yaniv Romano.
\newblock Model-{X} sequential testing for conditional independence via testing by betting.
\newblock In {\em International Conference on Artificial Intelligence and Statistics}, pages 2054--2086, 2023.

\bibitem{grunwald2023anytime}
Peter Gr{\"u}nwald, Alexander Henzi, and Tyron Lardy.
\newblock Anytime-valid tests of conditional independence under model-{X}.
\newblock {\em Journal of the American Statistical Association}, pages 1--12, 2023.

\bibitem{fedorova2012plug}
Valentina Fedorova, Alex Gammerman, Ilia Nouretdinov, and Vladimir Vovk.
\newblock Plug-in martingales for testing exchangeability on-line.
\newblock In {\em Proceedings of the International Conference on Machine Learning}, pages 923--930, 2012.

\bibitem{vovk2022testing}
Vladimir Vovk, Alexander Gammerman, and Glenn Shafer.
\newblock Testing exchangeability.
\newblock In {\em Algorithmic Learning in a Random World}, pages 227--263. Springer, 2022.

\bibitem{duan2022interactive}
Boyan Duan, Aaditya Ramdas, and Larry Wasserman.
\newblock Interactive rank testing by betting.
\newblock In {\em Conference on Causal Learning and Reasoning}, pages 201--235, 2022.

\bibitem{ramdas2022testing}
Aaditya Ramdas, Johannes Ruf, Martin Larsson, and Wouter~M Koolen.
\newblock Testing exchangeability: Fork-convexity, supermartingales and e-processes.
\newblock {\em International Journal of Approximate Reasoning}, 141:83--109, 2022.

\bibitem{saha2024testing}
Aytijhya Saha and Aaditya Ramdas.
\newblock Testing exchangeability by pairwise betting.
\newblock In {\em International Conference on Artificial Intelligence and Statistics}, pages 4915--4923, 2024.

\bibitem{grunwald2020safe}
Peter Gr{\"u}nwald, Rianne de~Heide, and Wouter~M Koolen.
\newblock Safe testing.
\newblock In {\em IEEE Information Theory and Applications Workshop (ITA)}, pages 1--54, 2020.

\bibitem{shafer2011test}
Glenn Shafer, Alexander Shen, Nikolai Vereshchagin, and Vladimir Vovk.
\newblock Test martingales, bayes factors and p-values.
\newblock {\em Statistical Science}, 26(1):84, 2011.

\bibitem{vovk2021values}
Vladimir Vovk and Ruodu Wang.
\newblock E-values: Calibration, combination and applications.
\newblock {\em The Annals of Statistics}, 49(3):1736--1754, 2021.

\bibitem{ramdas2022game}
Aaditya Ramdas, Peter Gr{\"u}nwald, Vladimir Vovk, and Glenn Shafer.
\newblock Game-theoretic statistics and safe anytime-valid inference.
\newblock {\em Statistical Science}, 38(4):576--601, 2023.

\bibitem{howard2021time}
Steven~R Howard, Aaditya Ramdas, Jon McAuliffe, and Jasjeet Sekhon.
\newblock Time-uniform, nonparametric, nonasymptotic confidence sequences.
\newblock {\em The Annals of Statistics}, 49(2), 2021.

\bibitem{waudby2023estimating}
Ian Waudby-Smith and Aaditya Ramdas.
\newblock Estimating means of bounded random variables by betting.
\newblock {\em Journal of the Royal Statistical Society Series B: Statistical Methodology}, 2023.

\bibitem{grunwald2023posterior}
Peter~D Gr{\"u}nwald.
\newblock The e-posterior.
\newblock {\em Philosophical Transactions of the Royal Society A}, 381(2247), 2023.

\bibitem{perez2022statistics}
Muriel~Felipe P{\'e}rez-Ortiz, Tyron Lardy, Rianne de~Heide, and Peter Gr{\"u}nwald.
\newblock E-statistics, group invariance and anytime valid testing.
\newblock {\em arXiv preprint arXiv:2208.07610}, 2022.

\bibitem{koolen2022log}
Wouter~M Koolen and Peter Gr{\"u}nwald.
\newblock Log-optimal anytime-valid e-values.
\newblock {\em International Journal of Approximate Reasoning}, 141:69--82, 2022.

\bibitem{vovk2023confidence}
Vladimir Vovk and Ruodu Wang.
\newblock Confidence and discoveries with e-values.
\newblock {\em Statistical Science}, 38(2):329--354, 2023.

\bibitem{shekhar2023reducing}
Shubhanshu Shekhar and Aaditya Ramdas.
\newblock Reducing sequential change detection to sequential estimation.
\newblock {\em arXiv preprint arXiv:2309.09111}, 2023.

\bibitem{shin2022detectors}
Jaehyeok Shin, Aaditya Ramdas, and Alessandro Rinaldo.
\newblock E-detectors: A nonparametric framework for sequential change detection.
\newblock {\em The New England Journal of Statistics in Data Science}, pages 1--32, 2023.

\bibitem{vovk2021testing}
Vladimir Vovk.
\newblock Testing randomness online.
\newblock {\em Statistical Science}, 36(4):595--611, 2021.

\bibitem{pmlr-v204-eliades23a}
Charalambos Eliades and Harris Papadopoulos.
\newblock A conformal martingales ensemble approach for addressing concept drift.
\newblock In {\em Conformal and Probabilistic Prediction with Applications}, volume 204, pages 328--346. PMLR, 2023.

\bibitem{vovk2020testing}
Vladimir Vovk.
\newblock Testing for concept shift online.
\newblock {\em arXiv preprint arXiv:2012.14246}, 2020.

\bibitem{dai2021testing}
Liang Dai and Mohamed-Rafik Bouguelia.
\newblock Testing exchangeability with martingale for change-point detection.
\newblock {\em International Journal of Ambient Computing and Intelligence (IJACI)}, 12(2):1--20, 2021.

\bibitem{vovk2021retrain}
Vladimir Vovk, Ivan Petej, Ilia Nouretdinov, Ernst Ahlberg, Lars Carlsson, and Alex Gammerman.
\newblock Retrain or not retrain: Conformal test martingales for change-point detection.
\newblock In {\em Conformal and Probabilistic Prediction and Applications}, pages 191--210. PMLR, 2021.

\bibitem{levy_pit}
Paul Lévy.
\newblock Théorie de l’addition de variables aléatoires. second edition 1954. (gauthier-villars, paris).
\newblock {\em The Mathematical Gazette}, 39, 1955.

\bibitem{wang2022continual}
Qin Wang, Olga Fink, Luc Van~Gool, and Dengxin Dai.
\newblock Continual test-time domain adaptation.
\newblock In {\em Proceedings of the IEEE/CVF Conference on Computer Vision and Pattern Recognition}, pages 7201--7211, 2022.

\bibitem{yuan2023robust}
Longhui Yuan, Binhui Xie, and Shuang Li.
\newblock Robust test-time adaptation in dynamic scenarios.
\newblock In {\em Proceedings of the IEEE/CVF Conference on Computer Vision and Pattern Recognition}, pages 15922--15932, 2023.

\bibitem{wang2023survey}
Zixin Wang, Yadan Luo, Liang Zheng, Zhuoxiao Chen, Sen Wang, and Zi~Huang.
\newblock In search of lost online test-time adaptation: {A} survey.
\newblock {\em arXiv preprint arXiv:2310.20199}, 2023.

\bibitem{guo2017calibration}
Chuan Guo, Geoff Pleiss, Yu~Sun, and Kilian~Q Weinberger.
\newblock On calibration of modern neural networks.
\newblock In {\em International Conference on Machine Learning}, pages 1321--1330. PMLR, 2017.

\bibitem{gong2022note}
Taesik Gong, Jongheon Jeong, Taewon Kim, Yewon Kim, Jinwoo Shin, and Sung-Ju Lee.
\newblock Note: Robust continual test-time adaptation against temporal correlation.
\newblock {\em Advances in Neural Information Processing Systems}, 35:27253--27266, 2022.

\bibitem{podkopaev2021distribution}
Aleksandr Podkopaev and Aaditya Ramdas.
\newblock Distribution-free uncertainty quantification for classification under label shift.
\newblock In {\em Uncertainty in artificial intelligence}, pages 844--853. PMLR, 2021.

\bibitem{zhou2023ods}
Zhi Zhou, Lan-Zhe Guo, Lin-Han Jia, Dingchu Zhang, and Yu-Feng Li.
\newblock Ods: Test-time adaptation in the presence of open-world data shift.
\newblock In {\em International Conference on Machine Learning}, pages 42574--42588. PMLR, 2023.

\bibitem{marsden2024universal}
Robert~A Marsden, Mario D{\"o}bler, and Bin Yang.
\newblock Universal test-time adaptation through weight ensembling, diversity weighting, and prior correction.
\newblock In {\em Proceedings of the IEEE/CVF Winter Conference on Applications of Computer Vision}, pages 2555--2565, 2024.

\bibitem{lu2023metatsallisentropy}
Menglong Lu, Zhen Huang, Zhiliang Tian, Yunxiang Zhao, Xuanyu Fei, and Dongsheng Li.
\newblock Meta-tsallis-entropy minimization: a new self-training approach for domain adaptation on text classification.
\newblock In {\em Proceedings of the International Joint Conference on Artificial Intelligence}, pages 5159--5169, 2023.

\bibitem{liu2021cycle}
Hong Liu, Jianmin Wang, and Mingsheng Long.
\newblock Cycle self-training for domain adaptation.
\newblock {\em Advances in Neural Information Processing Systems}, 34:22968--22981, 2021.

\bibitem{goyal2022testtime}
Sachin Goyal, Mingjie Sun, Aditi Raghunathan, and J~Zico Kolter.
\newblock Test time adaptation via conjugate pseudo-labels.
\newblock {\em Advances in Neural Information Processing Systems}, 35:6204--6218, 2022.

\bibitem{lee2013pseudo}
Dong-Hyun Lee et~al.
\newblock Pseudo-label: The simple and efficient semi-supervised learning method for deep neural networks.
\newblock In {\em Workshop on challenges in representation learning, ICML}, volume~3, page 896, 2013.

\bibitem{wang2023understanding}
Jun-Kun Wang and Andre Wibisono.
\newblock Towards understanding {GD} with hard and conjugate pseudo-labels for test-time adaptation.
\newblock In {\em International Conference on Learning Representations}, 2022.

\bibitem{schneider2020improving}
Steffen Schneider, Evgenia Rusak, Luisa Eck, Oliver Bringmann, Wieland Brendel, and Matthias Bethge.
\newblock Improving robustness against common corruptions by covariate shift adaptation.
\newblock {\em Advances in neural information processing systems}, 33:11539--11551, 2020.

\bibitem{sun2020test}
Yu~Sun, Xiaolong Wang, Zhuang Liu, John Miller, Alexei Efros, and Moritz Hardt.
\newblock Test-time training with self-supervision for generalization under distribution shifts.
\newblock In {\em International conference on machine learning}, pages 9229--9248. PMLR, 2020.

\bibitem{chen2021representation}
Xinyang Chen, Sinan Wang, Jianmin Wang, and Mingsheng Long.
\newblock Representation subspace distance for domain adaptation regression.
\newblock In {\em ICML}, pages 1749--1759, 2021.

\bibitem{borgwardt2006integrating}
Karsten~M Borgwardt, Arthur Gretton, Malte~J Rasch, Hans-Peter Kriegel, Bernhard Sch{\"o}lkopf, and Alex~J Smola.
\newblock Integrating structured biological data by kernel maximum mean discrepancy.
\newblock {\em Bioinformatics}, 22(14):e49--e57, 2006.

\bibitem{bhatnagar2023improved}
Aadyot Bhatnagar, Huan Wang, Caiming Xiong, and Yu~Bai.
\newblock Improved online conformal prediction via strongly adaptive online learning.
\newblock In {\em Proceedings of the International Conference on Machine Learning}, pages 2337--2363. PMLR, 2023.

\bibitem{panaretos2019statistical}
Victor~M Panaretos and Yoav Zemel.
\newblock Statistical aspects of {Wasserstein} distances.
\newblock {\em Annual review of statistics and its application}, 6:405--431, 2019.

\bibitem{ville19391ere}
Jean Ville.
\newblock {\em 1ere these: Etude critique de la notion de collectif; 2eme these: La transformation de Laplace}.
\newblock PhD thesis, Gauthier-Villars \& Cie, 1939.

\end{thebibliography}

\newpage

\appendix

\section{Additional related work: test-time adaptation}
\label{appdx:related_works}
As discussed in the main manuscript, entropy minimization is known to be effective for test time adaptation. However, the works in \cite{niu2022efficient,niu2022towards} demonstrate that samples with high entropy loss can lead to noisy or overly aggressive updates of model parameters. To address this issue, \cite{niu2022efficient,niu2022towards} filter out high-entropy samples, and \cite{niu2022towards} also employs gradient clipping to stabilize self-training. These algorithmic modifications emphasize the importance of controlling the minimization of the entropy loss, which greatly aligns with the goal of our work. In our framework, we also use entropy as a guiding force for self-training. However, instead of directly minimizing the entropy, our approach focuses on matching the distribution of the entropy losses at test time with that of the source domain. Notably, our method is versatile and can accommodate alternative choices beyond entropy, such as Tsallis entropy~\cite{lu2023metatsallisentropy,liu2021cycle} or cross-entropy evaluated with a pseudo-label~\cite{lee2013pseudo,goyal2022testtime,wang2023understanding} in place of the unknown true label. We leave the exploration of these alternative options for future work.

Another concern in test-time adaptation frameworks is the continuous learning mechanism, which often leads to performance degradation on in-distribution data. To address this challenge, \texttt{EATA} \cite{niu2022efficient} introduces an anti-forgetting strategy that optimizes the model by focusing on the reliability of samples and incorporates a weight regularizer to further improve stability. In our work, instead of relying on weight regularization, we preserve in-distribution performance through a ``no-harm'' approach, ensuring minimal model updates where no distribution shifts have occurred.

A crucial aspect of self-training is the selection of model parameters $\omega$ to update. A widely adopted practice is to update the parameters of the normalization layers. This constraint plays a vital role in mitigating overfitting, which is imperative to prevent the model from collapsing and making trivial predictions. The \texttt{TENT} \cite{wang2020tent} approach demonstrates that minimizing entropy by modifying only the batch normalization parameters can significantly enhance out-of-distribution performance. However,
\texttt{TENT} requires working with large batches, which can limit the ability to handle mixed-distribution shifts within a single batch---for instance, a batch containing a subset of blurred images, another subset of noisy images, and so on. To alleviate this issue, the \texttt{SAR} method suggests updating the group or layer normalization parameters, unlocking the use of smaller batch sizes. In turn, this adjustment enhances the model's adaptivity under mixed-distribution shifts. In our work, we follow this line of research and update the layer normalization parameters; however, we optimize a completely different loss function, aiming to match the distributions of the source and target entropies. While there are works that suggest matching between the source and target distributions \cite{schneider2020improving,sun2020test,chen2021representation}, this alignment is often affected by the batch size. To stress this point, one might consider using an out-of-the-box distributional matching loss function (e.g., Maximum Mean Discrepancy~\cite{borgwardt2006integrating}) to align the source and target entropy distributions, however, this approach requires large batches to obtain an effective estimation. Our approach accumulates the evidence for distribution shift in an online fashion, allowing us to use small batches, even of size one as we do in our experiments.

The work in \cite{wang2022continual} tackles the challenge of test-time adaptation where a pre-trained model must adjust to a continuously changing target domain during inference, without access to original training data. This approach, named \texttt{CoTTA}, utilizes weight-averaged and augmentation-averaged pseudo-labels to improve label accuracy and reduce error accumulation. To alleviate catastrophic forgetting, \texttt{CoTTA} intermittently reverts some neurons to their initial states. In contrast with \texttt{CoTTA}, our method operates under the assumption of no access to transformations/augmentations during testing. Employing such augmentations may further improve the performance of \texttt{POEM}, and we leave this exploration for future work.

\section{Learning the betting variable online}
\label{appdx:online}

In this section we present and analyze our online approach to learn the betting variable $\epsilon_j$, which relies on the SF-OGD algorithm \cite{orabona2018scale}. The complete algorithm is presented in Algorithm~\ref{alg:eps_sfogd}.

In the following analysis, we assume that the betting variable that attains the fastest growth of the capital in hindsight, after observing $u_\tau$, is in the range $[-D, D] \subset [-2, 2]$. We denote this variable by $E_\tau \in \{-D, +D\}$, which can be computed via a simple closed-form expression, given by \begin{equation}
\label{eq:E_tau}
    E_\tau = D \cdot \text{sign} (u_\tau - 0.5).
\end{equation} With this in place, we define the clip-aware loss function, which we will use to formulate the update rule for $\epsilon_\tau$ that maximizes the capital: 

\begin{equation}
\label{eq:sf-ogd-loss}
    L(E_\tau, \epsilon_\tau) = -\begin{cases} \log(1+E_\tau(u_\tau - 0.5)) \ & \text{if} \ \ E_\tau \cdot \epsilon_\tau > 0 \ \ \text{and} \ \ |\epsilon_\tau| > D \\ \log(1+\epsilon_\tau(u_\tau - 0.5)) 
    & \text{otherwise}.
    \end{cases}
\end{equation}
In plain words, when $\epsilon_\tau$ is out of range but has the same sign as $E_\tau$, we clip the betting variable with the maximal value allowed ($D$ or $-D$) that would increase the wealth. Otherwise, the loss is equal to the log of the bet, obtained with $\epsilon_\tau$.

At each step $\tau$ of the algorithm, we first predict the value of $\epsilon_\tau$, then observe $u_\tau$, which allows us to compute $E_\tau$. Therefore, after observing $u_\tau$ we can compute the (sub)gradient of \eqref{eq:sf-ogd-loss}:
\begin{equation}
\label{eq:nabla_log_bet}
    \nabla_\epsilon L(E_\tau, \epsilon_\tau) = -\begin{cases} 0 \ & \text{if} \ \ E_\tau \cdot \epsilon_\tau > 0 \ \ \text{and} \ \ |\epsilon_\tau| > D  \\ 
    (u_\tau - 0.5)/(1+\epsilon_\tau(u_\tau - 0.5)) & \text{otherwise}.
    \end{cases}
\end{equation}
Armed with $\nabla_\epsilon L(E_\tau, \epsilon_\tau)$ we are ready to perform the SF-OGD step, formulated as:
\begin{equation}
\label{eq:eps_update}
    \epsilon_{\tau+1} = \epsilon_{\tau} - \gamma \frac{\nabla_\epsilon L(E_\tau, \epsilon_{\tau})}{\sqrt{\sum_{t=1}^\tau (\nabla_\epsilon L_t(E_t, \epsilon_t))^2 }},
\end{equation}
where $\gamma>0$ is a learning rate. Lemma \ref{lemma:epsilon_bound} below states that $\epsilon_\tau$ is indeed bounded.
\begin{lemma}
    \label{lemma:epsilon_bound}
        The SF-OGD algorithm with a learning rate $0 < \gamma < 2 - D$ and initialization $\epsilon_1 = [-D-\gamma, \ D + \gamma]$ satisfies $\epsilon_\tau \in [-D-\gamma, \ D + \gamma]$ for all $1 \leq \tau \leq j$.
\end{lemma}

Proof is in Appendix \ref{proof:lemma_epsilon_bound}. Following Lemma~\ref{lemma:epsilon_bound}, we conclude that we must set the learning rate $\gamma$ in the range $0<\gamma<2-D$ to ensure that $\epsilon_\tau$ is in the range $(-2 ,2)$. The latter is crucial to formulate a valid betting function \eqref{eq:unif_test_martingale}.

Building on the analysis of SF-OGD, the following proposition states that this algorithm achieves a regret bound on the loss in \eqref{eq:sf-ogd-loss}, where the regret function is defined as
\begin{equation}
\label{eq:sf-gd-reg}
    \text{Reg}(j) = \sum_{\tau=1}^j L(E_\tau,\epsilon_\tau) - \sup_{\epsilon^\star \in [-D, D]} \sum_{\tau=1}^j L(E_\tau, \epsilon^*).
\end{equation}
Above, $\epsilon^\star$ is the betting parameter that minimizes the loss in hindsight, over the time horizon $1\leq t \leq j$.

\begin{theorem}
\label{thm:sfogd_regret}
[Theorem 4 by Orabona and p\'{a}l \cite{orabona2018scale}; Proposition A.2 by Bhatnagar et al. \cite{bhatnagar2023improved}]
    Suppose that $\epsilon^\star \in [-D, D]$. Then, SF-OGD with $E_\tau \in \{-D,D\}$ for all $1 \leq \tau \leq j$, learning rate $0<\gamma<2-D$, and any initialization $\epsilon_1 \in [-D - \gamma,D + \gamma]$ achieves
    $$\textup{Reg}(t) \leq (\gamma + \frac{1}{2\gamma}(2D+\gamma)^2) \sqrt{\sum_{\tau=1}^t{(\nabla_\epsilon L(T_\tau,\epsilon_\tau))^2}} \leq \mathcal{O}\left(\frac{\sqrt{t}}{2-D-\gamma}\right),  \ \  \forall 1 \leq t \leq j.$$
\end{theorem}
Proof is in Appendix \ref{proof:thm_sfogd_regret}. The above result states that for any interval of size $1 \leq t \leq j$, the regret of SF-OGD defined in~\eqref{eq:sf-gd-reg}, is bounded by the square-root of the interval size $\sqrt{t}$, divided by the difference between the boundaries of the entire $\epsilon_\tau$ domain $[-2,2]$ and of the actual $\epsilon_\tau$ domain $[D-\gamma,D+\gamma]$.

\section{Proofs}
\label{appdx:proofs}
\subsection{Proof of Proposition \ref{prop:validity}}
\begin{proof}{\ }
    \label{proof:prop_validity}
    $S_j \geq 0$ for all $j \in \mathbb{N}$ since $u_j \in [0,1]$ and $\epsilon_j \in [-1,1]$ by definition. \\
    $\{S_j: j \in \mathbb{N}\}$ is a martingale under $\mathcal{H}_0$ since
        \begin{align*}
            \mathbb{E}_{\mathcal{H}_0}[S_j \mid S_1,...,S_{j-1}] &= S_{j-1} \cdot \mathbb{E}_{\mathcal{H}_0}[b(u_j) \mid S_1,...,S_{j-1}] \\ &= S_{j-1} \cdot (1 + \epsilon_j \cdot \mathbb{E}_{\mathcal{H}_0}[u_j-0.5 \mid S_1,...,S_{j-1}]) = S_{j-1},
        \end{align*}
    where the second transition holds since $\epsilon_j$ depends only on $\{S_1,...,S_{j-1}\}$, and the latter since $\mathbb{E}_{\mathcal{H}_0}[u_j-0.5 \mid S_1,...,S_{j-1}] = 0$.
\end{proof}

\subsection{Proof of Lemma \ref{lemma:epsilon_bound}}
\begin{proof}
    \label{proof:lemma_epsilon_bound}
    Leveraging the symmetry of the problem, it suffices to study the setting in which $\epsilon_\tau \geq 0$. Also, without loss of generality, we assume that $\nabla_\epsilon L_1(E_1, \epsilon_1) \neq 0$ for the first iteration of the algorithm; if this does not hold we can always remove these samples until reaching an observation with a non-zero gradient. To prove the result, we consider the following cases that can occur when optimizing~\eqref{eq:sf-ogd-loss}. 
    \begin{itemize}
        \item \textbf{Case 1}: $0 \leq \epsilon_\tau \leq D$. 
        \item \textbf{Case 2}: $\epsilon_\tau > D $ and $\epsilon_\tau \cdot E_\tau \geq 0$.
        \item \textbf{Case 3}: $\epsilon_\tau > D $ and $\epsilon_\tau \cdot E_\tau \leq 0$.
    \end{itemize}

We start by analyzing \textbf{Case 1}. Recall that $0\leq u_\tau \leq 1$ and that $E_\tau \in \{-D, D\}$. Now, following \eqref{eq:nabla_log_bet} we can conclude that the gradient of the loss $L(E_\tau, \epsilon_\tau)$ is bounded $\nabla_\epsilon L(E_\tau, \epsilon_\tau) \in [-\frac{1}{2 - D}, \frac{1}{2 - D}]$. By recalling the update rule in \eqref{eq:eps_update}, we get:
    \begin{equation}
    \label{eq:epsilon_gap}
        |\epsilon_{\tau+1} - \epsilon_{\tau}| = \gamma \left| \frac{\nabla_\epsilon L(E_\tau, \epsilon_{\tau})}{\sqrt{\sum_{t=1}^\tau (\nabla_\epsilon L_t(E_t, \epsilon_t))^2 }} \right| \leq \gamma.
    \end{equation}
With this in place, we can conclude that if $\epsilon_{\tau+1} \geq \epsilon_{\tau}$, then $$\epsilon_{\tau+1} \leq \epsilon_{\tau} + \gamma \leq D + \gamma.$$ Otherwise, $\epsilon_{\tau+1} \leq \epsilon_{\tau}$, then $$D \geq \epsilon_{\tau} \geq \epsilon_{\tau+1} \geq \epsilon_{\tau} - \gamma \geq - \gamma.$$
Above, we used the fact that $0 \leq \epsilon_\tau \leq D$ under \textbf{Case 1}. In sum, we showed that $-\gamma \leq \epsilon_{\tau+1} \leq D+\gamma$.

We now turn to analyze \textbf{Case 2}. Under the assumptions of this case $\nabla_\epsilon L(E_\tau, \epsilon_{\tau}) = 0$ and thus $\epsilon_{\tau+1} = \epsilon_{\tau}$, i.e., the value of the betting parameter will not be modified. To bound the value of $\epsilon_{\tau+1}$, we note that we can encounter $\epsilon_\tau >D$ as a result of updating the betting parameter in \textbf{Case 1}, but in this scenario, we already know that $-\gamma \leq \epsilon_\tau \leq D + \gamma$. We can also reach \textbf{Case 2} at the initialization, but then $\epsilon_1 <= D +\gamma$ and thus bounded. Lastly, another entry point to \textbf{Case 2} is from \textbf{Case 3}, however, we show below that the latter satisfies that $D-\gamma \leq \epsilon_\tau \leq D + \gamma$.

Lastly, we study \textbf{Case 3}, which can be reached from \textbf{Case 1} or \textbf{Case 2}. However, in \textbf{Case 2} $\epsilon_{\tau+1} = \epsilon_{\tau}$, so we can concentrate only on the scenario where \textbf{Case 3} is reached from \textbf{Case 1}, but we already showed that $\epsilon_t \leq D + \gamma$ in this case. Lastly, we can face \textbf{Case 3} when $\epsilon_1 > D$, however, it is bounded $\epsilon_1 <= D +\gamma$ by the Lemma's assumption. Hence, following \eqref{eq:nabla_log_bet}, the gradient is bounded $\nabla_\epsilon L_t(E_t, \epsilon_t) \in [-\frac{1}{2 - D - \gamma}, \frac{1}{2 - D - \gamma}]$. Further, the loss can be improved only by reducing the value of  $\epsilon_{\tau}$, and thus the SF-OGD step would result in $\epsilon_{\tau+1} \leq \epsilon_{\tau}$. This implies that $\epsilon_{\tau} - \epsilon_{\tau+1} \leq \gamma$, according to \eqref{eq:epsilon_gap}. In turn, by recalling that $D + \gamma \geq \epsilon_\tau > D$ , we conclude that
$$D+\gamma \geq \epsilon_{\tau} \geq \epsilon_{\tau+1} \geq \epsilon_{\tau} - \gamma > D - \gamma.$$ 

To summarize, the analysis of the cases above indicates that $\epsilon_\tau \in [-D-\gamma, \ D + \gamma]$ for all $1 \leq \tau \leq j$, as desired.
\end{proof}

\subsection{Proof of Theorem \ref{thm:sfogd_regret}}
\begin{proof}
    \label{proof:thm_sfogd_regret}
    Recall that we want to prove that 
    $$\textup{Reg}(t) \leq (\gamma + \frac{1}{2\gamma}(2D+\gamma)^2) \sqrt{\sum_{\tau=1}^t{(\nabla_\epsilon L(T_\tau,\epsilon_\tau))^2}} \leq \mathcal{O}\left(\frac{\sqrt{t}}{2-D-\gamma}\right),  \ \  \forall 1 \leq t \leq j.$$
    The second inequality holds by following the proof of Lemma~\ref{lemma:epsilon_bound}, showing that $\nabla_\epsilon L_t(E_t, \epsilon_t) \in [-\frac{1}{2 - D - \gamma}, \frac{1}{2 - D - \gamma}]$. The proof of the first inequality is a direct consequence of \cite[Theorem 4]{orabona2018scale} or \cite[Proposition A.2]{bhatnagar2023improved}, and thus omitted. Notable, we can directly invoke \cite[Theorem 4]{orabona2018scale} as (i) the loss function $L_\tau(\cdot) = L(E_\tau, \cdot)$ in \eqref{eq:sf-ogd-loss} is convex; and (ii) the betting variable $\epsilon_\tau \in [-D-\gamma, D + \gamma]$ is bounded for all $1\leq \tau \leq j$ due Lemma~\ref{lemma:epsilon_bound}.
\end{proof}

\subsection{Proof of Proposition \ref{prop:optimal_transport}}

\begin{proof}
    \label{proof:prop_optimal_transport}
    Since we assume that $F_s$ is invertible and $Z_j^t$ is continuous, we can conclude that $F_s$ is a smooth bijection function. This allows us to invoke \cite[Lemma 1]{vovk2021protectedreg}, which states that if $F_t^j$ is the CDF corresponding to the density function $dF_t^j$, and the mapping $F_s$ is a smooth bijection, then the CDF $Q^{\text{opt}}(u) = F_t^j(F_s^{-1}(u))$ as in \eqref{eq:true_likelihood_u}. With this in place, we can write
    $$
    \tilde{Z}^t_j =  F_s^{-1}(Q^{\text{opt}}(u_j)) = F_s^{-1}(F_t^j(F_s^{-1}(u_j))) = F_s^{-1}(F_t^j(F_s^{-1}(F_s(Z_j^t)))) =  F_s^{-1}(F_t^j(Z_j^t)),
    $$
    where the second equality holds due \cite[Lemma 1]{vovk2021protectedreg} and the third equality stems from the definition of $u_j$, being $u_j=F_s(Z_j^t)$. We conclude the proof by observing that the mapping $\tilde{Z}^t_j = F_s^{-1}(F_t^j(Z_j^t))$ is the optimal transport map from the target to the source distribution w.r.t. the Wasserstein distance. This is because $Z_j^t$ is a continuous, one-dimensional random variable, with an invertible CDF $F_s$~\cite{panaretos2019statistical}. 
\end{proof}

\section{Using martingales to detect distribution shifts}
\label{appdx:shift_detection_with_martingale}

\begin{figure}[ht]
  \centering
  \includegraphics[width = \linewidth]{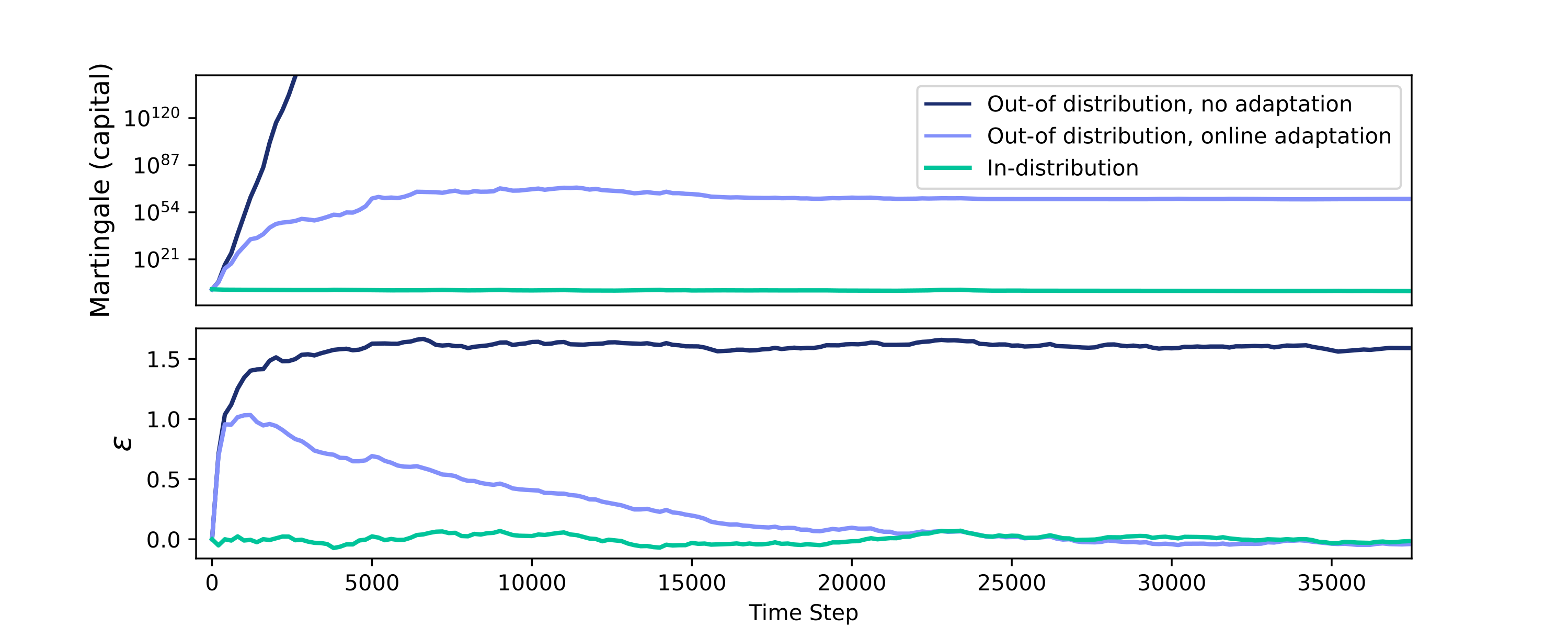}
  \caption{\textbf{Martingale behaviour with and without adaptation and on in-distribution data.} Visualization of three scenarios: (1) out-of-distribution data (ImageNet-C, brightness level 1) without adaptation, (2) the same out-of-distribution data with online adaptation, and (3) in-distribution data (ImageNet) all on ResNet50. The top panel shows the martingale value, that is, the accumulated capital (in $\log$ scale) over time, while the bottom panel shows the corresponding betting variable $\epsilon$.}
  \label{fig:appdx_martingale}
\end{figure}

Recall that we established in Proposition~\ref{prop:validity} the validity of the martingale defined in~\eqref{eq:unif_test_martingale}. In this section, we describe how a valid test martingale can be used to detect distribution shifts with a type-I error guarantee. This this end, consider a test martingale sequence denoted by $\{S_j: j\in\mathbb{N}_0\}$ under the null hypothesis $\mathcal{H}_0$ of no distribution shift \eqref{eq:null}. Ville's inequality~\cite{ville19391ere} plays a crucial role in bounding the probability of this martingale exceeding a specific threshold under $\mathcal{H}_0$. Specifically, for any value of $\alpha$ between 0 and 1, Ville's inequality states the following:
\begin{equation}
\label{eq:ville}
\mathbb{P}_{\mathcal{H}_0}\bigg(\exists j \ge 1: S_j \ge \frac{1}{\alpha}\bigg) \le \alpha \mathbb{E}_{\mathcal{H}_0}[S_0] = \alpha.
\end{equation}
Suppose we set a significance level $\alpha=0.01$. The above inequality states that under the assumption of no distribution shift, the martingale's value will exceed a threshold of ${1}/{\alpha}$ (in this case, $100$) with a probability of at most $\alpha=0.01$. This bound on the probability allows us to simultaneously control the type-I error across all time steps. This implies that we can declare that we face a distribution shift if the martingale value pass the threshold $1/\alpha$. For instance, if we set the threshold to $100$ or higher, the type-I error is guaranteed to be less than or equal to $0.01$.

Figure~\ref{fig:appdx_martingale} top panel offers a visual representation of the martingale's behavior under different scenarios; see Appendix~\ref{appdx:imp_details} for implementation details. As discussed above, under the null hypothesis of no distribution shift, the martingale is expected to remain lower than $1$ with high probability. This is precisely what we observe in the in-distribution data scenario---the martingale value remains under $1$. Here, we consider the ImageNet dataset, where we applied a pre-trained ResNet50 model on the ImageNet validation set.

In contrast, when the pre-trained model is applied to corrupted, out-of-distribution data (ImageNet-C), the martingale value deviates significantly from $1$, reaching levels of up to $10^{200}$. This substantial increase validates the presence of a distribution shift. Interestingly, Figure~\ref{fig:appdx_martingale} also reveals the effectiveness of our online adaptation. The adaptation process gradually limits the martingale value from growing compared to the non-adaptation case. Observe how the martingale of the adapted model eventually converges to a plateau. This visually demonstrates the success of adapting the model to the new distribution, making the target entropies statistically indistinguishable from the source data.

\section{Algorithms}
\label{sec:appdx_algs}
In this section, we present a series of algorithms essential to understanding and implementing our proposed methods. It is important to note that throughout these algorithms, we do not explicitly state each instance where lists or variables are updated; however, such updates are implicitly understood to occur during computations.

We use the ``\textbf{Assume}'' directive in our algorithms to outline which variables are accessible as global variables. These global variables are updated as part of the algorithms' operations but are not repeatedly declared within each algorithmic step. This approach is chosen to streamline the presentation and focus on the algorithmic logic rather than the mechanics of data handling.

\begin{algorithm}[H]
\caption{SF-OGD Step}\label{alg:eps_sfogd}
    \begin{algorithmic}[1]
        \Require $u_j \in[0, 1]$
        \Assume{{\color{white}{a}}
        \newline
        $\epsilon_j$: {\small\color{gray}last betting parameter's value}
        \newline
        $\{\nabla_\epsilon L(E_0, \epsilon_0), ..., \nabla_\epsilon L(E_{j -  1}, \epsilon_{j -  1})\}$}: {\small\color{gray}past gradient values of the log loss from \eqref{eq:sf-ogd-loss}}
        \newline $D$: {\small\color{gray}betting variable clip value}
        \newline
        $\gamma$: {\small\color{gray}SF-OGD learning rate}
        \newline{\color{white}Apply SF-OGD step}
        \State $E_j \gets D \cdot \text{sign}(u_j - 0.5)$ \Comment{Following~\eqref{eq:E_tau}}
        \If{$E_j \cdot \epsilon_j > 0 \ \text{and} \ |\epsilon_j| > D$}
        \State $\nabla_\epsilon L (E_j, \epsilon_j) \gets 0$ \Comment{Following \eqref{eq:nabla_log_bet}} 
        \Else
        \State $\nabla_\epsilon L (E_j, \epsilon_j) \gets - \frac{u_j - 0.5}{1 + \epsilon_j (u_j - 0.5)}$ \Comment{Following \eqref{eq:nabla_log_bet}}
        \EndIf
        \State $\epsilon_{j + 1} \gets \epsilon_j - \gamma \frac{\nabla_\epsilon L (E_j, \epsilon_j)}{\sqrt{\sum_{t=1}^j(\nabla_\epsilon L_t (E_t, \epsilon_t))^2}}$ \Comment{Following \eqref{eq:eps_update}}
        \State \Return $\epsilon_{j + 1}$
    \end{algorithmic}
\end{algorithm}

% Define a new length for comment alignment
\newlength{\commentindent}
\setlength{\commentindent}{.5\textwidth}

\makeatletter
% Define a new comment command for aligned comments
\renewcommand{\algorithmiccomment}[1]{%
  \unskip\hfill\makebox[\commentindent][l]{\(\triangleright\)~#1}\par%
}
\makeatother

\begin{algorithm}[H]
\caption{Protected Online Entropy Matching (\texttt{POEM})}\label{alg:pem_main_algo}
    \begin{algorithmic}[1]
        \Require {{\color{white}{a}}}
        \newline$\mathcal{D}^s = \{X_j^s\}_{j=1}^n$: {\small\color{gray}holdout data from source distribution}
        \newline$f_{\hat\theta}$: {\small\color{gray}pretrained model}
        \newline$D$: {\small\color{gray}last betting parameter's value}
        \newline$\gamma$: {\small\color{gray}SF-OGD learning rate}
        \newline$\eta$: {\small\color{gray}model learning rate}
        \newline$\lambda$: {\small\color{gray}entropy filter threshold, see~\eqref{eq:match_pp}}
        \newline {{\color{white}{a}}}
        \State \textbf{Init:} $\epsilon_1 = 0$
        , $\omega_1 \gets \mathbf{0}$ \Comment{We update only the network's norm. layers}
        \State {\color{gray}{Compute empirical CDF function of source entropies}}
        \For{$X_i^s$ in $\mathcal{D}^s$}
            \State $Z^s_i \gets \ell^{\text{ent}}(f_{\hat\theta}(X_i^s))$
        \EndFor
        \State Define: $\hat{F}_s(z) = \frac{1}{n}\sum_{i=1}^n \mathbbm{1}\{Z^s_i \leq z\}$ \Comment{Empirical CDF function} \label{alg_line:empirical_cdf}
        \State {\color{gray}{Online adaptation}}
        \State $j \gets 1$
        \While{Get a test sample $X_j^t$}
            \State $Z^t_j \gets \ell^{\text{ent}}(f_{\hat\theta + \omega_j}(X_j^t))$ \Comment{Compute test entropy}
            \State $u_j \gets \hat{F}_s(Z^t_j)$ \Comment{Apply probability integral transform}
            \State $b_j = 1 + \epsilon_j \cdot (u_j - 0.5)$ \Comment{Place a bet against the null \eqref{eq:null}} \label{alg_line:place_bet} 
            \State $\tilde{u}_j \gets \frac{1}{2} \epsilon_j u_j^2 + u_j\cdot(1 - \frac{\epsilon_j}{2})$ \Comment{Adapt $u_j$ according to \eqref{eq:eq_Q}} \label{alg_line:adapt_u} 
            \State $\tilde{Z}_j \gets \{\min z : \hat{F}_s(z) \geq \tilde{u}_j\}_{i=1}^n$ \Comment{Transport to source domain, $\hat{F}_s^{-1}(\tilde{u}_j)$} \label{alg_line:pseudo_inv_Fs}
            \State $\omega_{j+1} \gets \omega_j - \eta \nabla_{\omega} \ell^{\text{match++}}_{\lambda}({Z}^t_j,\tilde{Z}_j)$ \Comment{Update norm. layers' params. according to \eqref{eq:match_pp}}\label{alg_line:model_update} 
            \State $\epsilon_{j + 1} \gets$ Alg. \ref{alg:eps_sfogd} $(u_j)$ \Comment{Update the betting variable} \label{alg_line:sfogd} 
            \State $j \gets j + 1$
        \EndWhile        
    \end{algorithmic}
\end{algorithm}

\section{Supplementary experiments}
\label{appdx:exps_sup}
\label{appdx:imp_details}
\subsection{Synthetic experiment}

To evaluate the methods, we generate a synthetic dataset by generating two sets of points, each containing $2,500$ samples. Specifically, samples for the first set (class $-1$) are drawn from a normal distribution $\mathcal{N}(-1, 1)$, while samples for the second set (class $+1$) are drawn from $\mathcal{N}(1, 1)$. We use these $5,000$ samples to calculate the source CDF. Note that we do not use a training set, as we initialized the pre-trained model's parameter $\omega$ to the optimal value, which is $\omega=0$. 

We then create two test sets:
\begin{itemize}
    \item \textbf{In-distribution test data}, which consists of 10,000 samples per class, following the source distribution.
    \item \textbf{Out-of-distribution test data}, which consists of 10,000 samples per class, but with shifted distributions---the class $-1$ samples are drawn from $\mathcal{N}(0, 1)$ and the class $+1$ samples are drawn from $\mathcal{N}(2, 1)$. This represents a distributional shift of 1 unit in $X$.
\end{itemize}

We defined our model parameter as $\omega \in \mathbb{R}$. Two types of loss functions were employed:
\begin{itemize}
    \item Entropy loss $\ell^\text{ent}$ \eqref{eq:min_ent}, computed by evaluating the entropy of each point's prediction given the model parameter $\omega$.
    \item Matching loss $\ell^\text{match}$ \eqref{eq:match_loss}, computed using an optimal transport mapping, i.e., $F_s^{-1} \circ F_t$, to match the new set of target entropies derived from $f_{\omega}$ to the source entropies obtained from the holdout data.
\end{itemize}

\paragraph{Hyperparameters and training scheme} The optimization of each method was executed over 200 steps using a fixed learning rate of 5. Although lower learning rates were also effective, a higher learning rate was chosen to accelerate the demonstration. The optimization was performed with a batch size of 64. Each method's performance was evaluated at the end of this process. We clipped entropy minimization convergence if it went outside the bounds of our plot. In all experiments conducted in this paper, we choose the following set of hyperparameters, defined in Algorithm~\ref{alg:eps_sfogd}:
\begin{itemize}
    \item $D = 1.8$.
    \item $\gamma = \frac{1}{8 \cdot \sqrt{3}} \approx 0.0722$.
\end{itemize}

\subsection{ImageNet-C experiments}
\label{appdx:imgnet_imp_details}

\paragraph{Models} \rev{Our experiments utilize two pre-trained architectures: a Vision Transformer (ViT) with layer normalization (LN) and a ResNet50 with group normalization (GN). Both are pre-trained models from the \texttt{timm} library. We calibrate both models using temperature scaling, setting the temperature of ResNet50 to $T=0.90$ and $T=1.025$ for ViT. For \texttt{POEM} specifically, we implement an action delay of $100$ examples throughout the experiments in this paper. This delay allows the monitoring component of \texttt{POEM} to accumulate sufficient evidence before updating the model parameters, mitigating the influence of potentially noisy initial data.}

\paragraph{Data} \rev{We randomly sample 25\% of the examples from ImageNet validation set as an unlabelled holdout set. The corresponding corrupted examples are excluded from ImageNet-C to maintain the validity of the holdout data. All methods are evaluated only on the remaining 75\% samples. Each experiment is repeated 10 times with different holdout data splits and data selections, which is particularly crucial for experiments with small subsets of examples, such as the continual shifts experiments. Specifically for ViT, we modify the default preprocessing transforms offered by \texttt{SAR}, to the one used by the model during its training (see \url{https://huggingface.co/timm/vit_base_patch16_224.augreg2_in21k_ft_in1k} for the exact details). This adjustment is crucial to ensure proper estimation of the source CDF as well as ensure the model's performance on in-distribution data adheres to the one reported in Hugging Face.}

\paragraph{Code, hyperparameters, and learning scheme} \rev{We use the \texttt{SAR} \cite{niu2022towards} repository (available at \url{https://github.com/mr-eggplant/SAR}) for the ImageNet and ImageNet-C experiments. To ensure consistency with prior works, we adopt the hyperparameters, optimizers, and procedures provided within the \texttt{SAR} repository. This ensures that all baseline methods as well as \texttt{POEM} are run with the exact same settings. We also compare our method with \texttt{COTTA}\cite{wang2022continual} using the code provided by the authors, available at \url{https://github.com/qinenergy/cotta}.  In more detail:
\begin{itemize}
    \item For all methods except \texttt{COTTA}:
    \begin{itemize}
        \item The learning rate ($\eta$ in Algorithm~\ref{alg:pem_main_algo}) calculation follows these formulas:
        \begin{itemize}
            \item ViT: $\text{learning rate} = \left(\frac{0.001}{64}\right) \times \text{batch size}$
            \item ResNet50: $\text{learning rate} = \left(\frac{0.00025}{64}\right) \times \text{batch size} \times 2$
        \end{itemize}
        \item We use \texttt{SGD} optimizer with momentum of 0.9 for self-training.
        \item We use $\lambda = 0.40 \cdot \log{(1000)}$ (denoted by $E_0$ in ~\cite{niu2022efficient,niu2022towards}).
    \end{itemize}
    \item For \texttt{COTTA}:
    \begin{itemize}
        \item We tune the learning-rate search for each model (ResNet and ViT), ranging from $\frac{0.001}{64} \cdot i$ where $i\in [0.5, 1, 2, 4, 8]$, and select the best-performing value for each model on the continual setting with a corruption segment size of $1,000$. This process resulted in the following learning rate values.
        \begin{itemize}
            \item ViT $\text{learning rate} = \left(\frac{0.001}{64}\right) \times 4$
            \item ResNet50 $\text{learning rate} = \left(\frac{0.001}{64}\right) \times 2$
        \end{itemize}
        \item We use \texttt{Adam} optimizer with $\beta = (0.9, 0.999)$ and weight decay $0$ for self-training, as employed in the original work.
    \end{itemize}
\end{itemize}
A batch size of 1 is consistently used throughout all of the experiments, and, as mentioned in~Appendix \ref{appdx:imp_details}, we use $D=1.8$ and $\gamma=\frac{1}{8 \cdot \sqrt{3}} \approx 0.0722$ for our monitoring algorithm (see Algorithm~\ref{alg:eps_sfogd}). Lastly, since the empirical CDF in Algorithm~\ref{alg:pem_main_algo} (line~\ref{alg_line:empirical_cdf}) is calculated from a finite set of data points, it inherently creates a step function. In practice, we use linear interpolation to create a continuous function for better operation.
}

\paragraph{\rev{Hardware}} \rev{All experiments are conducted on our local server, equipped with 16 \texttt{NVIDIA A40 GPU - 49GB} GPUs, 192 \texttt{Intel(R) Xeon(R) Gold 6336Y} CPUs, and 1TB of RAM memory. Each experiment uses a single GPU and 8 CPUs.} \label{appdx:harware_imp_dets}

\subsubsection{Additional experiments: continual shifts}
\label{appdx:exps_cont}

\begin{figure}[ht]
  \centering
  \includegraphics[width = 400pt]{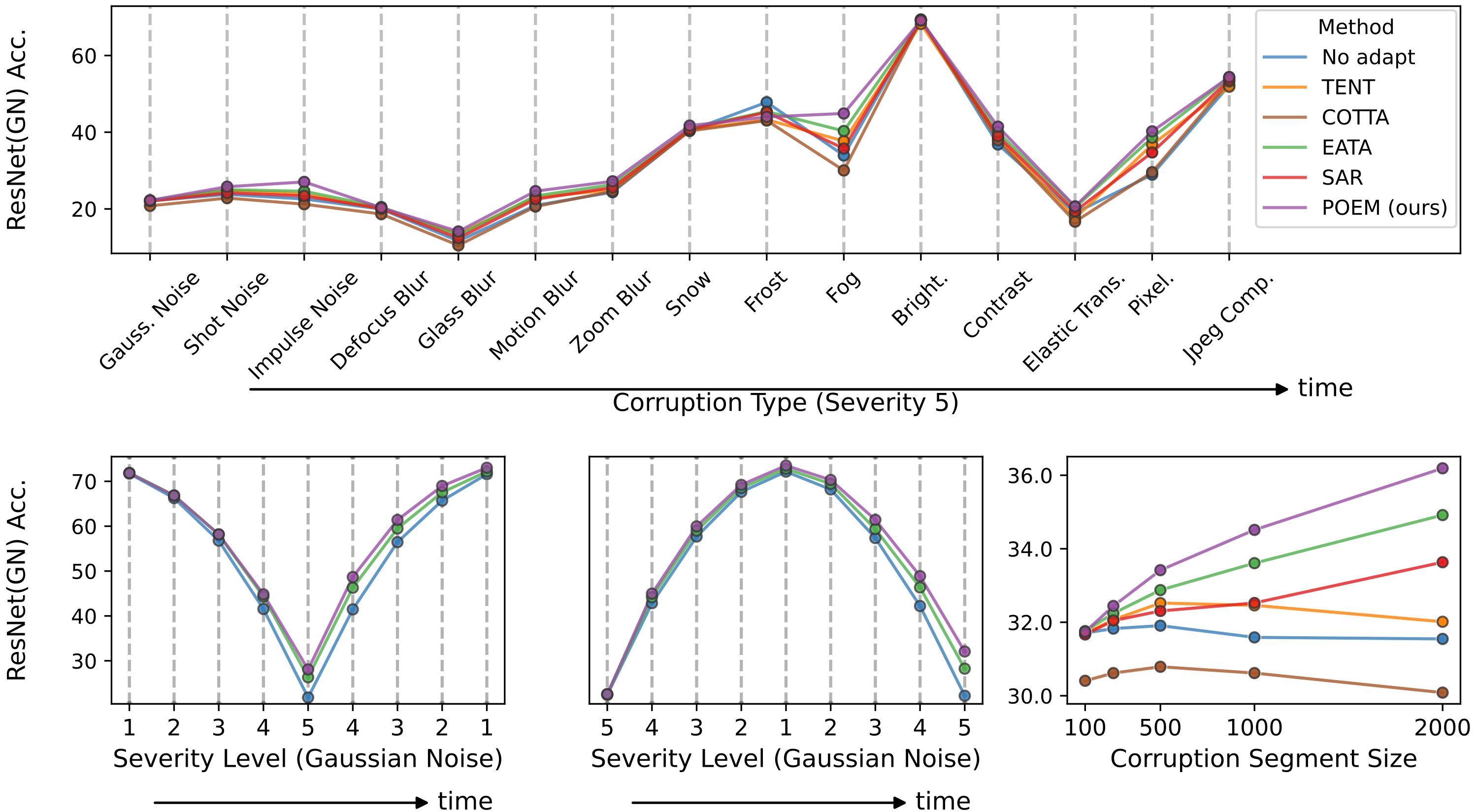}
  \caption{\textbf{Continual test-time adaptation on ImageNet-C with a ResNet model}. \textbf{Top:} Per-corruption accuracy with a corruption segment size of 1,000 examples. Results are obtained over 10 independent trials; error bars are tiny.  \textbf{Bottom left:} Severity shift---low~(1) to high~(5) and back to low. \textbf{Bottom center:} Severity shift---high~(5) to low~(1) and back to high. \textbf{Bottom right:} Mean accuracy under continual corruptions as a function of the corruption segment size.}
  \label{fig:appdx_cont_resnet}
\end{figure}

\paragraph{\rev{Supplementary details for the continual shifts experiments presented in the main manuscript}} \rev{For the corruption shift experiment, we used corruption segments sizes in the range of 100, 250, 500, 1000, and 2000. Apart from the bottom right panels of Figures~\ref{fig:cont_main} and~\ref{fig:appdx_cont_resnet}, the results are obtained using corruption segment size of 1,000 exclusively.}

\paragraph{\rev{Experiments with a ResNet (GN) model}} 
\rev{Herein, we repeat the same experiments from Section \ref{sec:exps} of the main manuscript to evaluate our proposed method within a continual shift setting, but now focus on a ResNet50 (GN) model. The results are summarized in Figure~\ref{fig:appdx_cont_resnet}, showing a similar trend to that obtained with the ViT model, albeit accuracies that are closer to the baseline methods. We note that the baseline accuracy of ResNet is far below that of ViT. The bottom right panel of Figure~\ref{fig:appdx_cont_resnet} demonstrates \texttt{POEM}'s efficiency in adapting to fast-changing distributions with only a few examples. While the advantage of \texttt{POEM} is less clear with small corruption segments in ResNet, it becomes evident as early as segment size 500. Observe the bottom right panel of Figure~\ref{fig:appdx_cont_resnet}. At corruption segment size of 2,000, \texttt{POEM} surpasses the best baseline method's accuracy (\texttt{EATA}) by 1.27\%. \texttt{POEM} also achieves 4.64\% increase compared to the original model (no adaptation).}

\subsubsection{Additional experiments: single shift}
\label{appdx:exps_tta}

\renewcommand{\arraystretch}{1.25}{
\setlength{\tabcolsep}{2pt}
\begin{table}[h!t]

\centering
\caption{\textbf{Summary of performance metrics of adaptation methods on the ImageNet-C dataset}, evaluated for each type of corruption at severity level 5. The results are evaluated on 10 independent experiments, conducted for each method and corruption type; standard error is presented.}
\label{tab:all_corruptions}
\resizebox{\linewidth}{!}{\begin{tabular}{llccccccccccccccc}
\toprule
Corruption &  &      Gauss. &          Shot &       Impulse &        Defocus &          Glass &         Motion &           Zoom &                Snow &               Frost &                 Fog &             Bright. &            Contrast &      Elastic &              Pixel. &          Jpeg \\  &          &         Noise &         Noise &         Noise &           Blur &           Blur &           Blur &           Blur &                     &                     &                     &                     &                     &         Trans. &                     &         Comp. \\
Method &                     &                     &                     &                     &                     &                     &                     &                     &                     &                     &                     &                     &                     &                     &                     \\
\midrule
\multirow{6}{*}{ResNet50 (GN)} & No adapt &            $22.11$ &            $23.05$ &            $22.04$ &            $\mathbf{19.82}$ &            $11.44$ &            $21.44$ &            $25.01$ &            $40.30$ &            $\mathbf{47.00}$ &            $33.98$ &            $68.82$ &            $36.26$ &            $18.55$ &            $29.24$ &            $52.59$ \\
             & \texttt{TENT} &  $12.22_{\pm 1.5}$ &  $14.50_{\pm 1.4}$ &  $15.80_{\pm 1.5}$ &  $14.49_{\pm 0.1}$ &   $6.36_{\pm 0.9}$ &  $20.62_{\pm 0.1}$ &  $20.59_{\pm 0.3}$ &  $22.37_{\pm 0.4}$ &  $27.90_{\pm 0.4}$ &   $1.88_{\pm 0.0}$ &  $70.26_{\pm 0.0}$ &  $42.65_{\pm 0.1}$ &   $8.10_{\pm 0.1}$ &  $49.65_{\pm 0.1}$ &  $54.55_{\pm 0.1}$ \\
             & \texttt{COTTA} &   $5.72_{\pm 0.1}$ &   $7.60_{\pm 0.1}$ &   $5.02_{\pm 0.1}$ &  $10.72_{\pm 0.5}$ &   $3.52_{\pm 0.0}$ &  $20.02_{\pm 0.2}$ &  $24.65_{\pm 0.1}$ &  $35.78_{\pm 0.2}$ &  $42.09_{\pm 0.1}$ &   $4.45_{\pm 0.1}$ &  $\mathbf{72.01_{\pm 0.1}}$ &  $34.92_{\pm 0.5}$ &  $16.03_{\pm 0.1}$ &  $19.97_{\pm 0.8}$ &  $55.96_{\pm 0.1}$ \\
             & \texttt{EATA} &  $31.22_{\pm 0.1}$ &  $33.07_{\pm 0.1}$ &  $32.51_{\pm 0.1}$ &  $18.41_{\pm 0.1}$ &  $18.61_{\pm 0.1}$ &  $29.41_{\pm 0.1}$ &  $30.11_{\pm 0.1}$ &  $\mathbf{45.11_{\pm 0.1}}$ &  $45.02_{\pm 0.1}$ &  $\mathbf{47.30_{\pm 0.1}}$ &  $70.76_{\pm 0.0}$ &  $45.06_{\pm 0.1}$ &  $\mathbf{28.18_{\pm 0.2}}$ &  $48.66_{\pm 0.1}$ &  $55.95_{\pm 0.1}$ \\
             & \texttt{SAR} &  $31.37_{\pm 0.1}$ &  $33.88_{\pm 0.1}$ &  $32.32_{\pm 0.1}$ &  $18.87_{\pm 0.1}$ &  $17.72_{\pm 0.1}$ &  $30.34_{\pm 0.0}$ &  $32.00_{\pm 0.1}$ &  $40.99_{\pm 0.5}$ &  $\mathbf{45.31_{\pm 0.0}}$ &  $21.33_{\pm 5.2}$ &  $\mathbf{72.01_{\pm 0.1}}$ &  $45.69_{\pm 0.1}$ &  $11.48_{\pm 0.1}$ &  $50.43_{\pm 0.1}$ &  $56.44_{\pm 0.0}$ \\
             & \texttt{POEM} (ours) &  $\mathbf{39.90_{\pm 0.1}}$ &  $\mathbf{42.16_{\pm 0.1}}$ &  $\mathbf{41.03_{\pm 0.1}}$ &  $18.86_{\pm 0.4}$ &  $\mathbf{22.14_{\pm 0.1}}$ &  $\mathbf{38.01_{\pm 0.1}}$ &  $\mathbf{36.16_{\pm 2.3}}$ &  $21.59_{\pm 0.7}$ &  $42.72_{\pm 2.6}$ &  $35.63_{\pm 7.5}$ &  $71.94_{\pm 0.1}$ &  $\mathbf{50.43_{\pm 0.1}}$ &   $9.06_{\pm 0.2}$ &  $\mathbf{55.85_{\pm 0.1}}$ &  $\mathbf{57.94_{\pm 0.0}}$ \\
\cline{1-17}
\multirow{6}{*}{ViT (LN)} & No adapt &            $49.73$ &            $50.27$ &            $50.03$ &            $42.72$ &            $34.43$ &            $50.57$ &            $44.79$ &            $56.61$ &            $52.31$ &            $56.63$ &            $75.86$ &            $31.93$ &            $46.89$ &            $65.53$ &            $66.39$ \\
             & \texttt{TENT} &  $57.70_{\pm 0.0}$ &  $58.80_{\pm 0.0}$ &  $58.67_{\pm 0.1}$ &  $57.28_{\pm 0.0}$ &  $53.18_{\pm 0.1}$ &  $60.47_{\pm 0.1}$ &  $56.87_{\pm 0.1}$ &  $64.93_{\pm 0.1}$ &  $52.78_{\pm 2.0}$ &  $68.39_{\pm 0.1}$ &  $78.42_{\pm 0.0}$ &  $61.86_{\pm 0.1}$ &  $61.31_{\pm 0.0}$ &  $72.16_{\pm 0.0}$ &  $70.29_{\pm 0.0}$ \\
             & \texttt{COTTA} &  $53.17_{\pm 0.4}$ &  $54.72_{\pm 0.3}$ &  $54.84_{\pm 0.3}$ &  $45.08_{\pm 1.1}$ &  $44.37_{\pm 0.3}$ &  $59.42_{\pm 0.2}$ &  $53.12_{\pm 0.1}$ &  $57.80_{\pm 0.9}$ &  $49.44_{\pm 0.5}$ &  $56.95_{\pm 1.1}$ &  $79.39_{\pm 0.0}$ &  $17.30_{\pm 1.5}$ &  $59.92_{\pm 0.2}$ &  $73.16_{\pm 0.2}$ &  $71.46_{\pm 0.1}$ \\
             & \texttt{EATA} &  $58.73_{\pm 0.0}$ &  $60.03_{\pm 0.1}$ &  $59.88_{\pm 0.1}$ &  $58.04_{\pm 0.0}$ &  $54.86_{\pm 0.1}$ &  $61.14_{\pm 0.0}$ &  $57.68_{\pm 0.1}$ &  $66.49_{\pm 0.1}$ &  $65.35_{\pm 0.0}$ &  $69.08_{\pm 0.1}$ &  $79.30_{\pm 0.0}$ &  $62.28_{\pm 0.1}$ &  $63.88_{\pm 0.1}$ &  $73.31_{\pm 0.1}$ &  $72.02_{\pm 0.0}$ \\
             & \texttt{SAR} &  $58.77_{\pm 0.0}$ &  $59.97_{\pm 0.0}$ &  $59.85_{\pm 0.1}$ &  $57.89_{\pm 0.0}$ &  $54.32_{\pm 0.1}$ &  $61.58_{\pm 0.0}$ &  $58.19_{\pm 0.1}$ &  $66.61_{\pm 0.1}$ &  $64.92_{\pm 0.1}$ &  $69.07_{\pm 0.1}$ &  $78.70_{\pm 0.1}$ &  $61.55_{\pm 0.1}$ &  $64.22_{\pm 0.1}$ &  $73.31_{\pm 0.0}$ &  $71.53_{\pm 0.1}$ \\
             & \texttt{POEM} (ours) &  $\mathbf{60.94_{\pm 0.0}}$ &  $\mathbf{62.60_{\pm 0.0}}$ &  $\mathbf{62.47_{\pm 0.1}}$ &  $\mathbf{60.08_{\pm 0.1}}$ &  $\mathbf{60.66_{\pm 0.1}}$ &  $\mathbf{65.23_{\pm 0.1}}$ &  $\mathbf{63.41_{\pm 0.0}}$ &  $\mathbf{70.05_{\pm 0.0}}$ &  $\mathbf{68.57_{\pm 0.1}}$ &  $\mathbf{73.39_{\pm 0.1}}$ &  $\mathbf{79.51_{\pm 0.0}}$ &  $\mathbf{63.99_{\pm 0.4}}$ &  $\mathbf{70.60_{\pm 0.0}}$ &  $\mathbf{75.45_{\pm 0.0}}$ &  $\mathbf{73.43_{\pm 0.1}}$ \\
\bottomrule
\end{tabular}}
\end{table}
}

Table~\ref{tab:tta_mean_results} summarizes baseline methods' accuracy on ImageNet-C for ViT and ResNet models. Table~\ref{tab:all_corruptions} offers a more detailed breakdown of the results in Table~\ref{tab:tta_mean_results}, showing accuracy for each corruption type.

\subsubsection{Additional experiments: in-distribution and the behavior of $\epsilon$}
\label{appdx:exps_in_dist}

\renewcommand{\arraystretch}{1.25}
\begin{table}[t]
\centering
\caption{Results of adaptation on in-distribution ImageNet data.}
\label{tab:in_dist_performance}
\resizebox{0.75\linewidth}{!}{
\begin{tabular}{llcccc}
\toprule
             &     &       Top-1 acc. &       Top-5 acc. &       Empirical cal. err. & $\|\omega\|_F^2$ \\
Model & Method &                   &                   &                  &                    \\
\midrule
\multirow{5}{*}{ResNet50 (GN)} & No adapt &  $79.95_{\pm 0.04}$ &  $94.94_{\pm 0.02}$ &  $0.0243_{\pm 0.00}$ &   $0.00_{\pm 0.00}$ \\
\cline{2-6} % Adjusted to start from the 2nd column to the 6th
             & \texttt{TENT} &  $79.63_{\pm 0.05}$ &  $94.78_{\pm 0.02}$ &  $0.0894_{\pm 0.00}$ &   $5.90_{\pm 0.01}$ \\
             & \texttt{COTTA} &  $79.87_{\pm 0.04}$ &  $94.83_{\pm 0.02}$ &  $0.0341_{\pm 0.00}$ &  $29.08_{\pm 0.02}$ \\
             & \texttt{EATA} &  $79.91_{\pm 0.04}$ &  $94.89_{\pm 0.02}$ &  $0.0709_{\pm 0.00}$ &  $10.53_{\pm 0.05}$ \\
             & \texttt{SAR} &  $79.97_{\pm 0.05}$ &  $94.93_{\pm 0.02}$ &  $0.0250_{\pm 0.00}$ &   $8.14_{\pm 0.01}$ \\
             & \texttt{POEM} (ours) &  $79.95_{\pm 0.04}$ &  $94.93_{\pm 0.02}$ &  $\mathbf{0.0243_{\pm 0.00}}$ &   $\mathbf{0.17_{\pm 0.03}}$ \\
\midrule % Or use \cline{2-6} if you prefer a partial line
\multirow{5}{*}{ViT (LN)} & No adapt &  $84.52_{\pm 0.03}$ &  $97.30_{\pm 0.02}$ &  $0.0099_{\pm 0.00}$ &   $0.00_{\pm 0.00}$ \\
\cline{2-6}
             & \texttt{TENT} &  $84.42_{\pm 0.04}$ &  $97.30_{\pm 0.02}$ &  $0.0135_{\pm 0.00}$ &   $4.35_{\pm 0.00}$ \\
             & \texttt{COTTA} &  $84.47_{\pm 0.04}$ &  $97.35_{\pm 0.01}$ &  $0.0209_{\pm 0.00}$ &  $13.75_{\pm 0.02}$ \\
             & \texttt{EATA} &  $84.57_{\pm 0.04}$ &  $97.35_{\pm 0.02}$ &  $0.0141_{\pm 0.00}$ &   $6.70_{\pm 0.05}$ \\
             & \texttt{SAR} &  $84.52_{\pm 0.03}$ &  $97.32_{\pm 0.02}$ &  $0.0101_{\pm 0.00}$ &   $0.86_{\pm 0.20}$ \\
             & \texttt{POEM} (ours) &  $84.48_{\pm 0.03}$ &  $97.29_{\pm 0.02}$ &  $\mathbf{0.0098_{\pm 0.00}}$ &   $\mathbf{0.32_{\pm 0.05}}$ \\
\bottomrule
\end{tabular}
}
\end{table}

\begin{figure}[t]
  \centering
  \includegraphics[width = \linewidth]{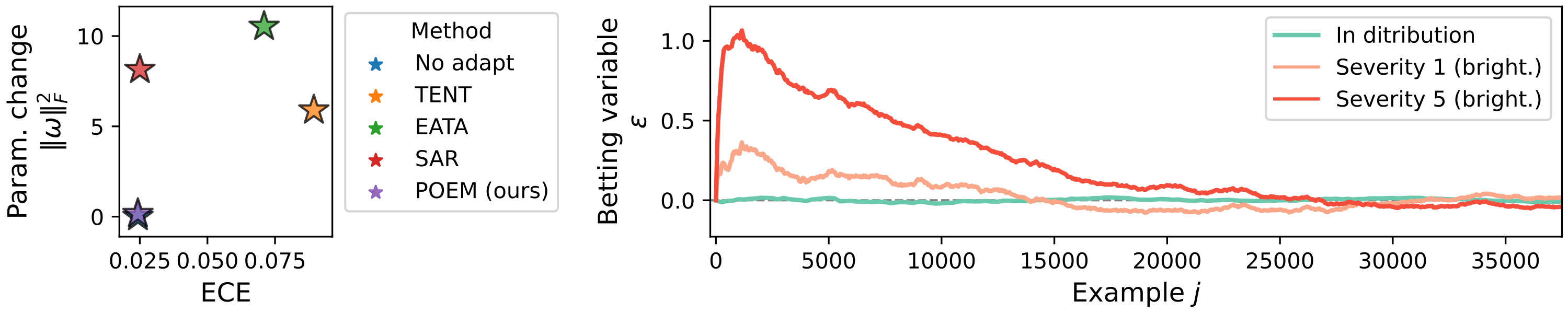}
  \caption{\textbf{In-distribution experiment on ImageNet (left panel)}: calibration error (ECE \cite{guo2017calibration}) versus $\|\omega\|_F^2$---a metric that evaluates the classifier's parameters deviation from the original ResNet50 model. Lower values on both axes are better. Results are averaged across 10 independent trials; standard errors and accuracy of each method are reported in Table~\ref{tab:in_dist_performance} in the appendix.  \textbf{The behavior of the betting parameter (right panel)}: the value of $\epsilon$ is presented as a function of time for both in- and out-of-distribution experiments (a single shift, two severity levels).}
  \label{fig:appdx_in_dist_resnet}
\end{figure}

\rev{Even though adapting on unseen in-distribution data from ImageNet attains similar accuracies for all baseline methods compared to the no-adapt approach (Table~\ref{tab:in_dist_performance}), \texttt{POEM} maintains this accuracy with the most minimal change to the original model's parameters $\omega$ and virtually unchanged ECE. This is demonstrated in the left panels of Figure~\ref{fig:in_dist_main} and Figure~\ref{fig:appdx_in_dist_resnet}. Specifically, as seen in Table~\ref{tab:in_dist_performance}, $\|\omega\|_F^2$ for \texttt{POEM} is merely $0.17$, significantly lower than other methods on ResNet50---over 30 times less change than the closest baseline method. This minimal change demonstrates \texttt{POEM}'s capability for controlled adaptation.} 

\begin{figure}[t]
  \centering
  \includegraphics[width = \linewidth]{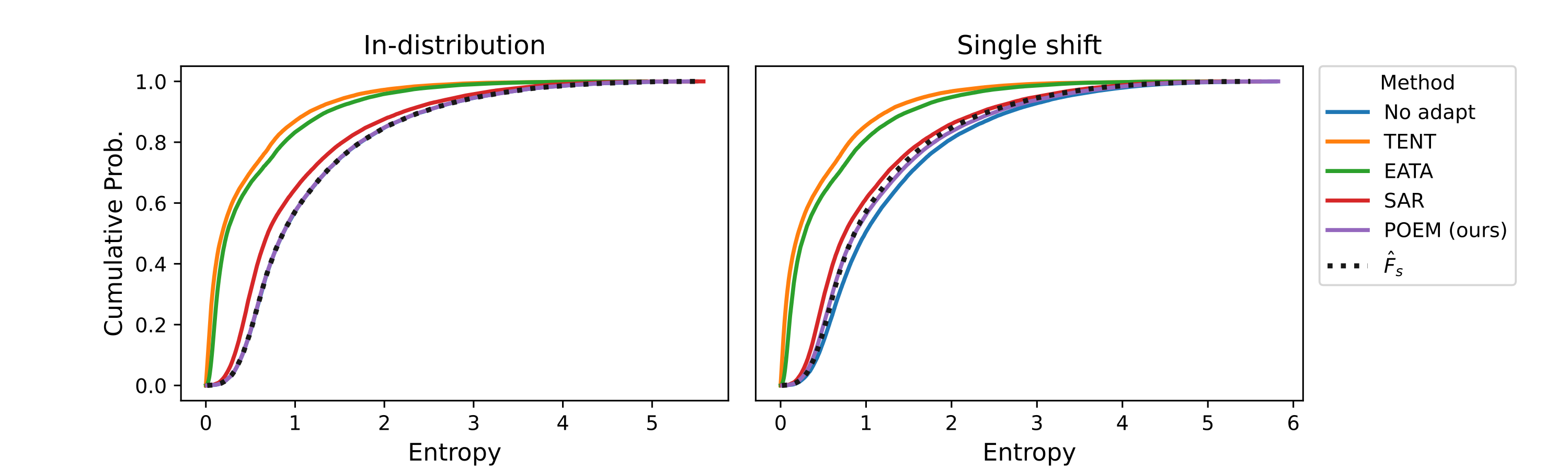}
    \caption{\textbf{Empirical test entropy CDF of each adaptation method, applied to in- and out-of-distribution ImageNet data.} The dotted black line represents the source CDF $\hat{F}_s$ obtained by applying the original ResNet50 model on test images from the source domain. The \textbf{left} panel shows how self-training on the validation set of ImageNet (in-distribution data) affects the entropy distribution of the model. The \textbf{right} panel repeats the experiment on out-of-distribution data from ImageNet-C with brightness corruption of severity level 1.}
  \label{fig:appdx_cdfs}
\end{figure}

\rev{Figure~\ref{fig:appdx_cdfs} further demonstrates the ``no-harm'' effect of \texttt{POEM} under in-distribution test data. The left panel shows the empirical CDF of the entropy values $\hat{F}_t$ of all baseline methods, indicating that these lead to overconfident predictions. This is in contrast to \texttt{POEM}, whose estimated target CDF closely aligns with the source distribution. Such a minimal deviation from the source distribution aligns with our previous findings---\texttt{POEM} tends to keep the model parameters intact when adaptation is unnecessary.}

\rev{When applying the model to out-of-distribution data (right panel of Figure~\ref{fig:appdx_cdfs}), the unadapted model appears slightly under-confident, as indicated by its corresponding CDF $\hat{F}_t$ being lower than $\hat{F}_s$. Here, \texttt{POEM} effectively restores the model's original confidence by adjusting its estimated CDF to closely match the source CDF. 
\texttt{SAR} achieves comparable results to \texttt{POEM}, but through a different strategy. It employs a restart mechanism that acts as a safeguard, activated when the entropy of the adapted model drops below a specific exponential moving average (EMA) threshold.}

\subsubsection{Ablation study}
\label{appdx:ablation}

\begin{table}
\centering
\caption{\textbf{Ablation study: the effect of} $\ell^{\text{match}}$ \textbf{compared to}  $\ell^{\text{match++}}$ \textbf{on the accuracy of \texttt{POEM}}. Results are presented for ImageNet-C and averaged over all 15 corruptions of severity level 5.}
\label{tab:ablation_avg}
\resizebox{0.50\linewidth}{!}{
\begin{tabular}{llll}
\toprule
Method     & $\ell^{\text{match}++}$ &  ResNet50 (GN) &  ViT (LN) \\
\midrule
No adapt & n/a &             31.46 &         51.65 \\
% \cline{1-4}
\texttt{POEM} & \xmark &             32.49 &         60.64 \\
     & \cmark &             38.90 &         67.36 \\
\bottomrule
\end{tabular}}
\end{table}

\rev{We assess the impact of $\ell^{\text{match++}}$~\eqref{eq:match_pp} compared to $\ell^{\text{match}}$~\eqref{eq:match_loss} on adaptation accuracy through isolated analysis. Table~\ref{tab:ablation_avg} shows that our basic match loss $\ell^{\text{match}}$ improves the test accuracy over the no-adapt baseline, even without filtering and weighting. This underscores the core strength of our approach. The enhanced $\ell^{\text{match}++}$ that incorporates sample filtering and weighting mechanisms further boosts performance.}

\subsection{CIFAR10-C and CIFAR100-C experiments}
\label{appdx:exps_cifar}

\paragraph{Data} \rev{CIFAR10-C and CIFAR100-C datasets are extensions of the original CIFAR10 and CIFAR100 test sets. These datasets consist of 15 different corrupted versions of the original CIFAR test images, with each corruption applied at 5 severity levels, mirroring the structure of ImageNet-C. In our experimental setup, we randomly select $25\%$ of the examples from original CIFAR test set to create unlabelled holdout sets of sizes 2,500 images. To preserve the integrity of the holdout set, we exclud the corresponding corrupted examples from CIFAR10-C and CIFAR100-C, respectively. All adaptation methods were applied on the remaining $75\%$ of the data, ensuring consistency across approaches, regardless of their holdout set requirements. We conduct each experiment for $10$ independent trials.}

\paragraph{Model} Our experiments utilize a pre-trained ResNet32 model with batch normalization (BN), obtained from \texttt{torch-hub} and available at \url{https://github.com/chenyaofo/pytorch-cifar-models}.

\paragraph{Methods, code, hyperparameters, and learning scheme} \rev{The pre-trained ResNet32 architecture includes batch normalization (BN) layers, which forces us to use a batch-size of 4 during self-training. This differs from the batch-size we used in the ImageNet experiments, which was equal to 1. In what follows, we compare \texttt{POEM} to \texttt{SAR}, \texttt{EATA}, and \texttt{TENT} only. We do not conduct experiments with \texttt{COTTA}, as our ImageNet experiments showed that \texttt{COTTA} performs inferiorly when the self-training batch size is small.
To ensure fair comparison, we perform a grid search for choosing the optimal learning rate $\eta$ for each method. The details of this sensitivity analysis are presented in the following section. In the case of \texttt{POEM}, we retained the hyperparameters of the monitoring tool as outlined in Section~\ref{appdx:imp_details}.}

\subsubsection{\rev{Continual shifts experiments}}
\begin{figure}[t]

  \centering
  \begin{subfigure}{\textwidth}
    \centering
    \includegraphics[width=1\textwidth]{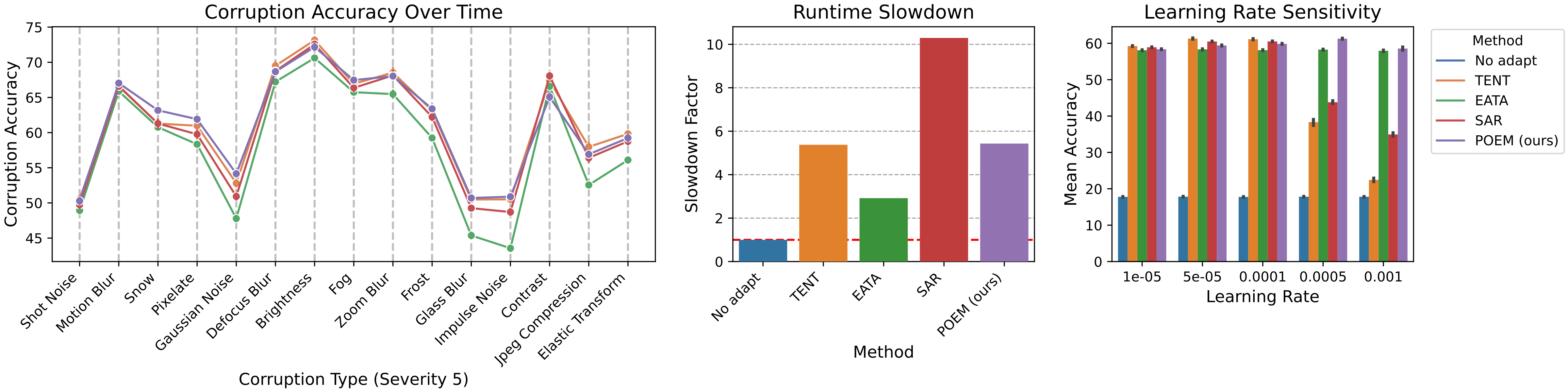}
  \end{subfigure}
  
  \vspace{0.2cm}
  
  \begin{subfigure}{\textwidth}
    \centering
    \includegraphics[width=1\textwidth]{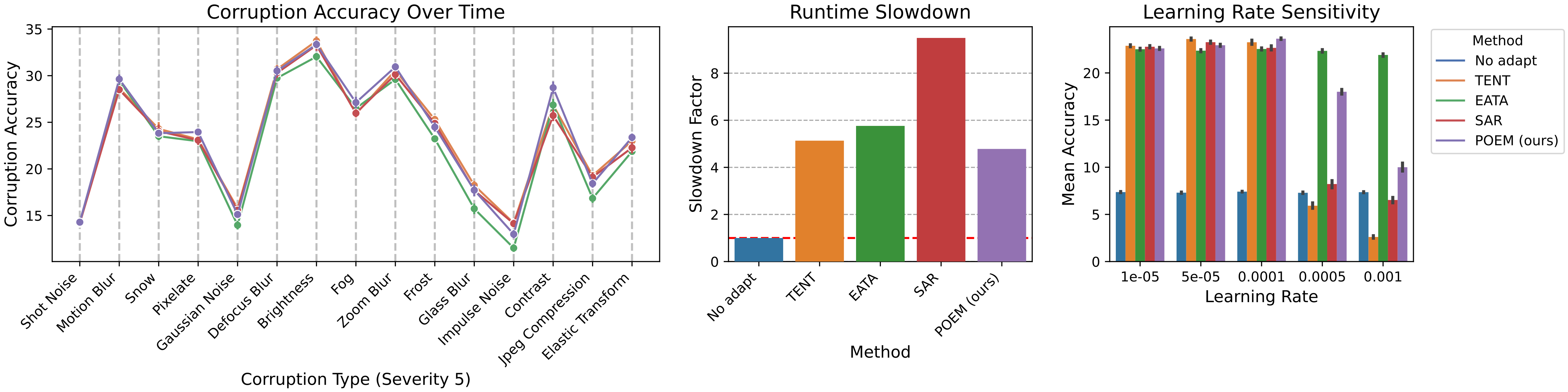}
  \end{subfigure}
  \caption{\rev{\textbf{Short-term adaptation performance on a test set of about 15,360 samples from CIFAR10-C (top) and CIFAR100-C (bottom) using a ResNet (BN) model.}
  \textbf{Left:} Continual test-time adaptation performance showing per-corruption accuracy with a corruption segment size of 1,024 examples of severity level 5. Results are averaged over 10 independent trials with error bars indicated. The `no adapt' baseline is not displayed as its mean accuracy is significantly lower than other methods (see the right panel); we omitted this method to enhance visualization of differences between the adaptation techniques. \textbf{Middle:} Runtime slowdown, defined as the runtime of test-time adaptation divided by the runtime of the source (no-adapt) model; lower is better. \textbf{Right:} Learning rate sensitivity study, showing mean performance across the continual experiment. The results in the left panel are obtained with the best learning rate for each method.}}
  \label{fig:cifar-short}
\end{figure}

\begin{figure}[ht]

  \centering
  \begin{subfigure}{\textwidth}
    \centering
    \includegraphics[width=1\textwidth]{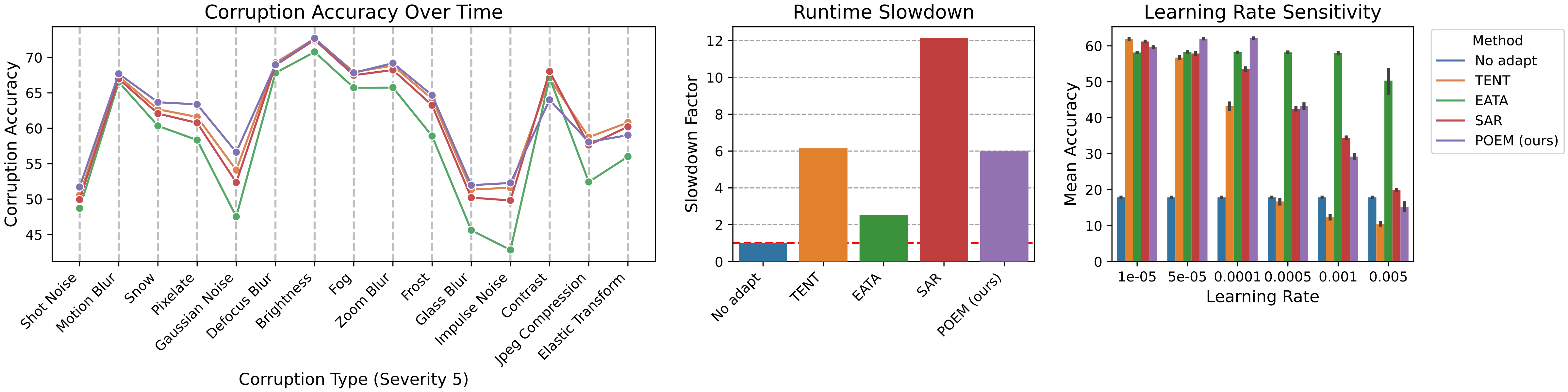}
  \end{subfigure}
  
  \vspace{0.2cm}
  
  \begin{subfigure}{\textwidth}
    \centering
    \includegraphics[width=1\textwidth]{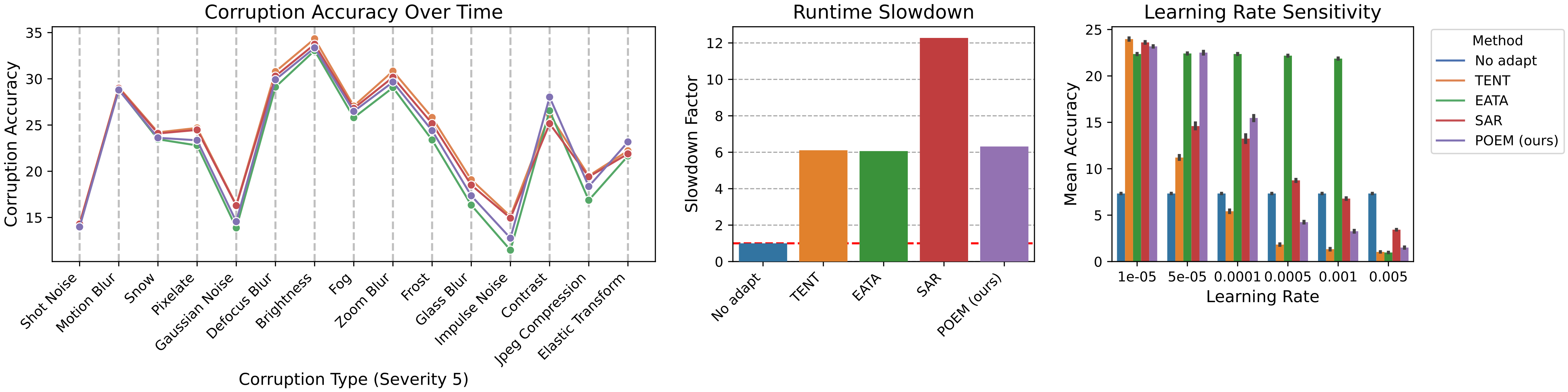}
  \end{subfigure}
  \caption{\rev{\textbf{Long-term adaptation performance on a test set of 112,500 samples from CIFAR10-C (top) and CIFAR100-C (bottom) using a ResNet (BN) model.}
  Continual test-time adaptation is applied with a corruption segment size of 7,500 examples of severity level 5. The other details are as in Figure~\ref{fig:cifar-short}.}}
  \label{fig:cifar-long}
\end{figure}

\paragraph{\rev{Short-term adaptation performance}} \rev{We focus on the continual setting where the corruption type is changing over time, akin to our ImageNet experiments. Each corruption type includes $1,024$ samples, resulting in a test set of approximately $15 \cdot 1,024 \approx 15,000$ samples. The results are summarized in Figure~\ref{fig:cifar-short}. Following that figure, one can see that our method is competitive and even outperforms baseline methods in terms of accuracy. Runtime comparisons (relative to the no-adapt model) are also presented, demonstrating that our method's complexity is similar to \texttt{TENT} and \texttt{EATA}, and lower than \texttt{SAR}. Additionally, the sensitivity study for the learning rate parameter $\eta$ reveals our method's robustness to this hyperparameter, particularly when compared to \texttt{SAR} and \texttt{TENT}.}

\paragraph{\rev{Long-term adaptation performance}} \rev{Here, we conduct a similar experiment, but on a larger test set containing 112,500 corrupted samples (15 versions of 7,500 images). The results, summarized in Figure~\ref{fig:cifar-long}, show that our proposed method is competitive with the baseline methods in terms of adaptation accuracy. Notably, our method's runtime is twice as fast as \texttt{SAR} and comparable to \texttt{EATA} and \texttt{TENT}. Unlike \texttt{SAR}, we do not employ a model-reset mechanism, nor do we use an anti-forgetting loss like \texttt{EATA}---yet our approach demonstrates robustness over the long term. In contrast, the right panel of Figure~\ref{fig:cifar-long} reveals that \texttt{TENT} is highly sensitive to the choice of learning rate.}

\subsection{Office-Home experiments}
\label{appdx:exps_officehome}

\paragraph{Data} \rev{The OfficeHome dataset consists of images from 4 domains: ``Art'', ``Clipart'', ``Product'', and ``Real World''. It contains a total of 15,588 images across 65 object categories. The ``Art'' domain has 2,427 images, ``Clipart'' has 4,365 images, ``Product'' has 4,439 images, and ``Real World'' has 4,357 images. We focus on adaptation from the ``Real World'' domain to the ``Art'', ``Clipart'', and ``Product'' domains. We chose this setup over a continual setting as it is deemed more natural for this dataset. Given the lack of a predefined data structure, we split the dataset into an $80\%$ training set from the ``Real World'' samples, with the remainder serving as validation and holdout sets for our method and \texttt{EATA}.}

\paragraph{Methods, model, hyperparameters, and learning scheme} \rev{We fine-tune the last layer of the ResNet50 with Group Normalization (GN) previously used in the ImageNet experiments. We fit the model on $80\%$ of the ``Real World'' examples for 25 epochs, with the best model saved based on performance on the remaining $20\%$. We use Adam optimizer with default PyTorch hyperparameters and set the learning rate to $5\cdot 10^{-5}$. Similar to the CIFAR experiments, we compare \texttt{POEM} to \texttt{SAR}, \texttt{EATA}, and \texttt{TENT} only; our ImageNet experiments showed that \texttt{COTTA} performs inferiorly when the self-training batch size is 1, which is used in this experiment as well. Learning rates are tuned for each method using a predefined grid, ensuring a fair comparison, similar to our CIFAR experiments. The hyperparameters of \texttt{POEM}'s monitoring tool are as specified in Section~\ref{appdx:imp_details}.}

\subsubsection{\rev{Single domain adaptation experiments}}

\begin{figure}[t]
  \centering
  \includegraphics[width=1\textwidth]{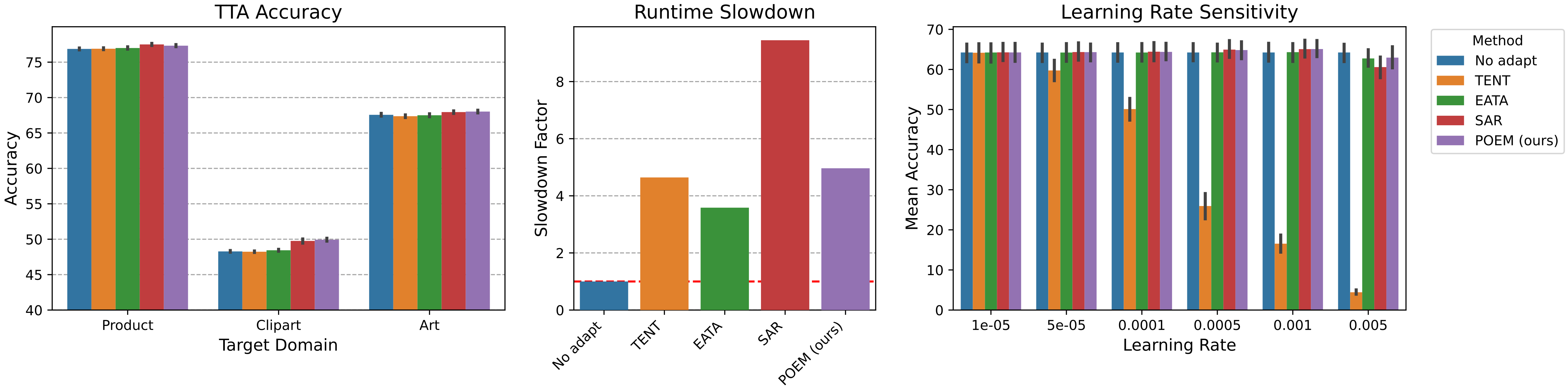}
  \caption{\rev{\textbf{Performance analysis of test-time adaptation methods on the Office-Home dataset using a ResNet (GN) model, pre-trained on ImageNet and fine-tuned on the ``Real World'' source domain.} \textbf{Left:} Test-time accuracy for adaptation from the source domain to three distinct target domains (``Art'', ``Clipart'', and ``Product''). Results are evaluated on the complete test dataset of each target domain and averaged across 10 independent trials, with error bars indicated. 
  \textbf{Middle:} Runtime slowdown comparison. \textbf{Right:} Learning rate sensitivity study, displaying the average accuracy across all three target domains.
   The results in the left panel are obtained using the best learning rate for each method.}}
  \label{fig:office_home}
\end{figure}

\rev{Test-time adaptation is applied to the entire test data of each target domain (''Art'', ''Clipart'' and ''Product''). Results are summarized in Figure~\ref{fig:office_home}. Overall, all the methods demonstrate modest accuracy gains compared to the `no-adapt' case. Our proposed method \texttt{POEM} slightly outperforms \texttt{TENT} and \texttt{EATA} in terms of accuracy, while achieving results comparable to \texttt{SAR}. In terms of computational efficiency, our method's runtime is on par with \texttt{TENT} and \texttt{EATA}, and notably faster than \texttt{SAR}. Regarding sensitivity to the choice of the learning rate, our approach displays superior robustness compared to \texttt{TENT} and \texttt{SAR}, and a similar robustness to that of \texttt{EATA}.}

\end{document}